\documentclass{article}
\usepackage{ijcai22}

\usepackage{times}
\usepackage{soul}
\usepackage{url}
\usepackage[hidelinks]{hyperref}
\usepackage[utf8]{inputenc}
\usepackage[small]{caption}
\usepackage{graphicx}
\usepackage{amsmath}
\usepackage{amsthm}
\usepackage{booktabs}
\newcommand{\citep}[1]{~\cite{#1}}
\newcommand{\citet}[1]{~\cite{#1}}

\usepackage{amsmath,amsfonts,bm}

\def\eqref#1{equation~\ref{#1}}

\def\1{\bm{1}}

\def\vtheta{{\bm{\theta}}}
\def\va{{\bm{a}}}

\def\vs{{\bm{s}}}

\def\vw{{\bm{w}}}
\def\vx{{\bm{x}}}
\def\vy{{\bm{y}}}
\def\vz{{\bm{z}}}

\def\mA{{\bm{A}}}

\def\mD{{\bm{D}}}

\def\mI{{\bm{I}}}

\def\mX{{\bm{X}}}
\def\mY{{\bm{Y}}}
\def\mZ{{\bm{Z}}}

\DeclareMathAlphabet{\mathsfit}{\encodingdefault}{\sfdefault}{m}{sl}
\SetMathAlphabet{\mathsfit}{bold}{\encodingdefault}{\sfdefault}{bx}{n}

\newcommand{\E}{\mathbb{E}}

\newcommand{\KL}{D_{\mathrm{KL}}}

\DeclareMathOperator*{\argmax}{arg\,max}
\DeclareMathOperator*{\argmin}{arg\,min}

\usepackage{amsfonts}
\usepackage{amsmath}
\usepackage{amssymb}

\usepackage{algorithm}
\usepackage{algpseudocode} 
\usepackage{algorithmicx}
\usepackage{mathtools}
\usepackage{xfrac} 

\usepackage{caption} 
\usepackage{subcaption}  
\usepackage{cleveref}
\usepackage{tcolorbox}
\usepackage[inline,shortlabels]{enumitem}
\usepackage{xcolor} 
\usepackage{hyperref}
\usepackage{booktabs}

\usepackage{titlesec}

\usepackage{enumitem}
\usepackage{bbm}
\setlist{leftmargin=2em}

\let\given\givenbase

\usepackage{amsthm}

\newcommand{\norm}[1]{\left\lVert #1 \right\rVert}

\newcommand{\lpz}{M}  

\makeatletter
\let\alignts@preamble\align@preamble
\patchcmd{\alignts@preamble}{\displaystyle}{\textstyle}{}{}
\patchcmd{\alignts@preamble}{\displaystyle}{\textstyle}{}{}

\def\alignts{\let\align@preamble\alignts@preamble\start@align\@ne\st@rredfalse\m@ne}

\makeatother

\makeatletter
\newcommand{\algmargin}{\the\ALG@thistlm}
\makeatother
\algnewcommand{\parState}[1]{\State%
    \parbox[t]{\dimexpr\linewidth-\algmargin}{\strut #1\strut}}

\usepackage{graphicx}
\usepackage[export]{adjustbox}
\usepackage{wrapfig}
\usepackage[font=small,labelfont=bf]{caption}

\newtheorem{proposition}{Proposition}[section]

\newcommand{\LL}{\mathcal{L}}

\newcommand{\blue}[1]{\textcolor{blue}{#1}}

\title{Diverse Imitation Learning via Self-Organizing Generative Models}

\author{
    Author Name
    \affiliations
    Affiliation
    \emails
    pcchair@ijcai-22.org
}

\author{
Arash Vahabpour$^1$
\and
Tianyi Wang$^1$\and
Qiujing Lu$^1$
\and
Omead Pooladzandi$^1$\and
Vwani Roychowdhury$^1$
\affiliations
$^1$University of California, Los Angeles (UCLA)
\emails
\{vahabpour, tianyiw, qiujing, opooladz, vwani\}@ucla.edu
}

\begin{document}

\maketitle

\begin{abstract}

Imitation learning is the task of replicating expert policy from demonstrations, without access to a reward function. This task becomes particularly challenging when the expert exhibits a mixture of behaviors.
Prior work has introduced latent variables to model variations of the expert policy. However, our experiments show that the existing works do not exhibit appropriate imitation of individual modes.
To tackle this problem, we adopt an encoder-free generative model for behavior cloning (BC) to accurately distinguish and imitate different modes. Then, we integrate it with GAIL to make the learning robust towards compounding errors at unseen states. We show that our method significantly outperforms the state of the art across multiple experiments.

\end{abstract}

\section{Introduction}
The goal of imitation learning is to learn to perform a task from expert trajectories without a reward signal. Towards this end, two approaches are studied in the literature. The first approach is behavior cloning (BC) \citep{pomerleau1991efficient}, which learns the mapping between individual expert states-action pairs in a supervised manner. It is well known that BC neglects long-range dynamics within a trajectory. This leads to problem of compounding error \citep{ross2010efficient,ross2011reduction}, that is, when a trained model visits states that are slightly different from those visited in expert demonstrations, the model takes erroneous actions, and the errors quickly accumulate as the policy unrolls. Therefore, BC can only succeed when training data is abundant. 

An alternative approach, known as inverse reinforcement learning (IRL) \citep{ziebart2008maximum}, recovers a reward function that makes the expert policy optimal. Such an approach often requires iterative calculation of the optimal policy with reinforcement learning (RL) in an inner loop, making it computationally expensive. 
The computational cost can be reduced by jointly optimizing a policy---via RL, and a cost function---via maximum causal entropy IRL \citep{ziebart2008maximum,ziebart2010modeling}. The resulting algorithm is called generative adversarial imitation learning (GAIL) \citep{ho2016generative}, and is closely connected to generative adversarial networks \citep{goodfellow2014generative}. 
GAIL has shown better promise than BC, both in sample efficiency in the number of expert trajectories, and in robustness towards unseen states.

Expert demonstrations of a particular task can display variations, even given the same initial state. This diversity can often stem from multiple human experts performing the same task in different ways. Alternatively, it might arise from pursuing multiple unlabeled objectives, e.g. a robotic arm moving towards different goals. In such scenarios, neither BC nor GAIL enjoys any explicit mechanism to capture different modes of behavior.
Particularly, BC averages out actions from different modes. On the other hand, GAIL learns a policy that captures only a subset of control behaviors, because adversarial training is prone to mode collapse \citep{li2017infogail,wang2017robust}.

Imitation of diverse behaviors is addressed in several prior works. \citet{fei2020triple,merel2017learning} use labels of expert modes in training. These works differ from ours in that we assume expert modes are unlabeled. InfoGAIL \citep{li2017infogail} and Intention-GAN \citep{hausman2017multi} augment the objective of GAIL with the mutual information between generated trajectories and the corresponding latent codes. This mutual information is evaluated with the help of a posterior network. \citet{wang2017robust} use a variational autoencoder (VAE) module to encode expert trajectories into a continuous latent variable. These latent codes are supplied as an additional input to both the policy and the discriminator of GAIL. Nevertheless, as we show in our experiments, both InfoGAIL and VAE-GAIL perform poorly in practice.

Multimodal imitation learning requires finding the correspondence between expert trajectories and a latent code. \citet{wang2017robust} suggest learning this correspondence using the encoder in a VAE. However, this imposes the difficulty of designing an isolated (as opposed to end-to-end) recurrent module to encode sequences of state-action pairs. Moreover, VAEs are mainly well suited for continuous latent variables. We propose to address these challenges by removing the recurrent encoder module, and instead directly search over samples of the latent variable.
\citet{vondrick2016anticipating,bojanowski2017optimizing,hoshen2019non} have recently considered similar encoder-free models. 

Prior work on encoder-free models has mainly focused on non-sequential data. Our work is the first to consider encoder-free models for sequential decision making, and in particular, multimodal imitation learning. We choose the name self-organizing generative model (SOG) for our core algorithm, indicating that it learns semantically smooth latent spaces. We present a unified framework for both discrete and continuous latent variables. In both cases, we analyse the relationship between our SOG algorithm and algorithms based on maximization of the marginal data likelihood, e.g. \citet{kingma2013auto}. Moreover, we propose an extension to our SOG algorithm that enables training with high dimensional latent spaces. Finally, we combine SOG---as a BC method---with GAIL, to get the best of both worlds. This attempt is inspired by the empirical study of \citet{jena2020augmenting}, where in a unimodal setting, a combination of BC with GAIL is shown to yield better performance in fewer training iterations.


In a nutshell, this work mainly focuses on imitation of diverse behaviors while distinguishing different modes. Our contributions are summarized as follows:
\begin{enumerate}
    \item We adopt SOG as an encoder-free generative model that can be used for multimodal BC. We show that SOG procedure is closely connected to maximizing the marginal likelihood of data. Besides, we introduce an extension of SOG that enables learning with high-dimensional latent spaces.
    \item We integrate SOG with GAIL to benefit from accuracy of SOG and robustness of GAIL at the same time. Thus, the resulting model (a) can generate trajectories that are faithful to the expert, (b) is robust to unseen states, (c) learns a latent variable that models variations of the expert policy, and (d) captures all modes successfully without mode collapse.
    \item Our empirical results show that our model considerably outperforms the baselines.
\end{enumerate}

\section{Background} \label{sec:background}
\subsection{Preliminaries}
We consider an infinite-horizon discounted Markov decision process (MDP) as a tuple $(\mathcal{S}, \mathcal{A}, P, r, \rho_0, \gamma)$. Namely, $\mathcal{S}$ denotes the state space, $\mathcal{A}$ denotes the action space, $P:\mathcal{S}\times\mathcal{A}\times\mathcal{S}\rightarrow\mathbb{R}$ represents the state transition probability distribution, $r:\mathcal{S}\times\mathcal{A}\rightarrow\mathbb{R}$ is the reward function, $\rho_0:\mathcal{S}\rightarrow \mathbb{R}_{\geq 0}$ represents the distribution over initial states, and $\gamma\in (0,1)$ is the reward discount factor. 

Let $\pi:\mathcal{S}\times\mathcal{A}\rightarrow\mathbb{R}_{\geq 0}$ denote a stochastic policy function. Demonstrations of the policy $\pi$ are sequences of state-action pairs, sampled as follows: $\vs_0\sim\rho_0$, $\va_t\sim\pi(\va_t|\vs_t)$, $\vs_{t+1}\sim P(\vs_{t+1}|\vs_t,\va_t)$. When states and actions generated by policy $\pi$, we define expected return as $\mathbb{E}_\pi\left[r(\vs,\va)\right]\coloneqq \mathbb{E}_\pi\left[\sum_{t=0}^\infty \gamma^t r(\vs_t,\va_t)\right]$.
\subsection{Imitation Learning}
The goal of imitation learning is to replicate the expert policy from demonstrations without access to the underlying reward function. Two approaches exist for this problem. The first approach is BC, which treats the state-action pairs as inputs and outputs of a policy function, and optimizes for the $L_2$ loss between the expert actions and the predicted actions. The second approach is GAIL, which optimizes the following objective:
\begin{equation} \label{eq:gail}
\begin{split}
\min_{\substack{\pi_{\boldsymbol\theta}}}\max_{\substack{D}} \,\{&\mathbb{E}_{\pi_{\boldsymbol\theta}}\left[\log D(\vs,\va)\right]\\
+&\mathbb{E}_{\pi_E}\left[\log \left(1-D(\vs,\va)\right)\right]\\
-&\lambda H(\pi_{\boldsymbol\theta})\}.
\end{split}
\end{equation}
In this equation, $\pi_{\boldsymbol\theta}$ denotes the policy function realized by a neural network with weights $\boldsymbol\theta$, and $\pi_E$ denotes the expert policy. Moreover, $D$ denotes the discriminative classifier that distinguishes between state-action pairs from the trajectories generated by $\pi_{\boldsymbol\theta}$ and $\pi_E$. Lastly, $H(\pi_{\boldsymbol\theta})\coloneqq\E_{\pi_{\boldsymbol\theta}}[-\log \pi_{\boldsymbol\theta}(\va|\vs)]$ represents the discounted causal entropy of the policy $\pi_{\boldsymbol\theta}$ \citep{bloem2014infinite}.
We provide additional details about optimization of \Cref{eq:gail} in \Cref{appx:gail-optim}. 
\subsection{Imitation of Diverse Expert Trajectories}
We model the variations in the expert policy with a latent variable $\vz$. In particular, $\vz$ is sampled from a discrete or continuous prior distribution $p(\vz)$ before each rollout, and specifies the mode of behavior. Therefore, we formalize the generative process of a multimodal expert policy as follows: $\vz\sim p(\vz),\vs_0\sim\rho_0,\va_t\sim\pi_E(\va_t|\vs_t,\vz),\vs_{t+1}\sim P(\vs_{t+1}|\va_t,\vs_t)$.

\section{Method} \label{sec:method}

In this section, we aim to propose an algorithm for imitation of diverse behaviors. We first introduce a generative model for learning the distribution of arbitrary datasets. Namely, let $\vx$ and $\vy$ respectively denote an input data point and a corresponding output data point. We assume that $\vy$ is generated through a two-stage random process:
\begin{enumerate}
\item A  latent variable $\vz$ (independent of $\vx$) is sampled from a prior distribution $p(\vz)$. This prior can be a given discrete distribution over $K$ categories with probability masses $\pi_1,\ldots,\pi_K$, or an arbitrary continuous distribution, e.g. a multivariate Gaussian $\displaystyle p(\vz)= \mathcal{N} ( \vz ; \mathbf{0}, \mI)$.
\item A value $\vy$ is sampled from a conditional distribution $p(\vy|\vz, \vx;f)$  of the following form:
\begin{equation} \label{eq:posterior}
    p(\vy|\vz,\vx;f) = \mathcal{N} ( \vy ; f(\vz,\vx), \sigma^2\mI),
\end{equation}
where $f$ maps $\vz$ and $\vx$ to the mean of the distribution, and $\sigma^2$ is the isotropic noise variance in the space of $\vy$.
\end{enumerate}

\subsection{The SOG Algorithm in General} \label{sec:general-sog}
Given a dataset $\mathcal{D}=\{(\vx_{i},\vy_{i})\}_{i=1}^N$, consisting of $N$ i.i.d samples generated by the process above, we design an algorithm  to estimate $\boldsymbol\theta$, such that the samples generated by the inferred model match the given data in distribution.
First, one needs to restrict the estimation of function $f$ within a family of parameterized functions, e.g. neural networks $f_{\boldsymbol\theta}$ with weight parameters $\boldsymbol\theta$. Next, starting with a random initialization of $\boldsymbol\theta$, an iterative algorithm is adopted: 
\begin{enumerate*}[(1)]
\item Given current $\boldsymbol\theta$, one computes estimates of the latent variables $\vz_i$, $i=1, \cdots, N$, corresponding to each data point. In a marginal likelihood framework, a posterior distribution for $\vz_i$ is estimated. In our SOG algorithm, we instead obtain point estimates $\vz_i = \vz_i^*$, where $\vz_i^* = \argmax_{\vz_i} p(\vy_i|\vz_i,\vx_i;\,\boldsymbol\theta)$.  If the latent variable is discrete, $\vz_i^*$ is found by an exhaustive search over $\vz\in\{1,\ldots,K\}$. In the continuous case, the optimal $\vz$ is searched over samples of the prior distribution $p(\vz)$.  \item Given the best latent codes, one takes a gradient step on $\boldsymbol\theta$ to increase the likelihood $p(\vy_i|\vz_i^*,\vx_i;\,\boldsymbol\theta)$.
\end{enumerate*}
Both steps decrease the same loss function, guaranteeing convergence.

\begin{algorithm}[H]
	\caption{Self-Organizing Generative Model (SOG)}
	\label{alg:sog}
	\begin{algorithmic}[1]
	    \State \textbf{Define}: Loss function $\mathcal{L}\left(\hat{\vy},\vy\right)\coloneqq ||\hat{\vy}-\vy||^2$
	    \State \textbf{Input:} Initial parameters of policy network $\boldsymbol\theta_0$
		\For {$epoch=1,2,\ldots$}
		     \For {$iteration=1,2,\ldots$}
		        \parState{Sample a minibatch of $N_\text{data}$ data points $ (\vx_i,\vy_i)$ from the dataset.}
		        \For {$i=1,2,\ldots,N_\text{data}$}
    		        \State Sample $N_z$ latent codes $ \vz_j \sim p(\vz)$. \label{alg:sog:latent-batch}
    	           \State Calculate $\vz_i\coloneqq\arg\min_{\vz_j}\mathcal{L}\left(f_{\boldsymbol\theta}(\vz_j,\vx_i), \vy_i\right)$. \label{alg:sog-best-z}
    	       \EndFor
    	       \parState{Calculate $ \mathcal{L}_{\mathrm{SOG}}=\sum_{i=1}^{N_\text{data}}\mathcal{L}\left(f_{\boldsymbol\theta}(\vz_i,\vx_i), \vy_i\right)$ and its gradients w.r.t. ${\boldsymbol\theta}$.} \label{alg:sog-loss}
    	       \parState{Update $f_{\boldsymbol\theta}$ by stochastic gradient descent on ${\boldsymbol\theta}$ to minimize $\mathcal L_{\mathrm{SOG}}$.}
		    \EndFor
		\EndFor
	\end{algorithmic}
\end{algorithm}
Formal relationships and equivalences between the SOG algorithm -- that iteratively computes a Maximum Likelihood Estimate (MLE) of $\vz$ for given $\boldsymbol\theta$ and then increases the likelihood of data computed at these MLE estimates of $\vz$ by updating the model parameters $\boldsymbol\theta$ -- and algorithms based on maximizing the marginalized likelihood -- that iteratively compute a posterior distribution of $\vz$ for given $\boldsymbol\theta$, and then increase the likelihood of data marginalized over the distribution of $\vz$ by updating the model parameters $\boldsymbol\theta$ -- are included in \Cref{appx:em,appx:continuous}.
\subsection{Adopting SOG for Multimodal Imitation Learning}
\label{sec:sog-imitation}
In the following, we introduce two approaches for multimodal imitation learning: 
\begin{enumerate*}[(1)]
    \item adopting SOG for multimodal BC, and
    \item combining multimodal BC with GAIL using SOG.
\end{enumerate*}
Due to space limits, the corresponding algorithms are presented in \Cref{appx:algorithms}.
\subsubsection*{Multimodal Behavior Cloning with SOG} We adopt \Cref{alg:sog} for learning the relationship between state-action pairs in a multimodal BC setting. We enforce an additional constraint that the latent code is shared across each trajectory. Hence, we propose \Cref{alg:sog-bc}.
\subsubsection*{Multimodal Combination of BC and GAIL using SOG}
Next, we introduce \Cref{alg:sog-gail}, where we combine SOG, as a means for BC, with GAIL to ensure robustness towards unseen states. This algorithm is inspired by \citet{jena2020augmenting}, which in a unimodal setting optimizes a weighted sum of BC loss and the GAIL ``surrogate'' loss (see \Cref{appx:gail-optim}).
\subsection{Further Properties of SOG} \label{sec:sog-properties}
\subsubsection*{Sample Complexity} Training procedure of \Cref{alg:sog}, given a high-dimensional continuous latent variable, requires a prohibitively large number of latent code samples (\cref{alg:sog:latent-batch} of the algorithm). In \Cref{appx:extended-sog}, we propose a computationally efficient extension of this algorithm suited for high dimensions of the latent space.

\subsubsection*{Self-Organization in the Latent Space} In \Cref{appx:lipschitz-bounds}, we show that latent codes corresponding to nearby data points get organized close to each other. In addition, visual results in  \Cref{appx:latent-interpret,appx:extended-sog} show that different regions of the latent space organize towards generating different modes of data. Similar properties have been studied in ``self-organizing maps'' \cite{bishop2006pattern}, without a deep network. Hence, we chose the phrase ``self-organization'' in naming of \Cref{alg:sog}.
\section{Experiments} \label{sec:experiments}
In our experiments, we demonstrate that our method can recover and distinguish all modes of expert behavior, and replicate each robustly and with high fidelity. We evaluate our models, SOG-BC and SOG-GAIL, against two multimodal imitation learning baselines: InfoGAIL \citep{li2017infogail} and VAE-GAIL \citep{wang2017robust}. 

\subsection{Experiment Setup} \label{sec:experimental-setup}
We adopt an experiment from \citet{li2017infogail} in which an agent moves freely at limited velocities in a 2D plane. In this experiment, the expert produces three distinct circle-like trajectories. In another series of experiments, we evaluate our model on several complex robotic locomotion tasks, simulated via the MuJoCo simulator \citep{todorov2012mujoco}. These experiments, borrowed from \citet{rakelly2019efficient}, include multimodal tasks with discrete or continuous modes. Discrete tasks include Ant-Fwd-Back, Ant-Dir-6, HalfCheetah-Fwd-Back, Humanoid-Dir-6, Walker2d-Vel-6, and Hopper-Vel-6. Additionally, we conduct two experiments with continuous modes, namely FetchReach and HalfCheetah-Vel. All these experiments are named after standard MuJoCo environments. The suffix ``Fwd-Back'' indicates that the modes correspond to forward or backward moving directions. Besides, the suffix ``Dir-6'' implies six moving directions, namely $k\cdot2\pi/6$ angles. On the other hand, ``Vel'' and ``Vel-6'' indicates that the expert selects velocities uniformly at random, resulting in different modes of behavior. The sets of expert velocities are listed in \Cref{appx:experiment-details}. Later, we refer to the environments Ant, HalfCheetah, Humanoid, Walker2d, and Hopper, as ``locomotion'' tasks.

The FetchReach experiment involves a simulated 7-DoF robotic arm \citep{plappert2018multi}. At the beginning of each episode, a target point is uniformly sampled from a cubic region. The expert controls the arm to reach the desired target within a tolerance range. 

We explain more details about our experiment setup in \Cref{appx:experiment-details}.

\subsection{Evaluation}
\begin{table*}[htp] 
\caption{Mean and standard deviation of the rewards for locomotion tasks.}
\label{tab:rewards}
\begin{center}
\resizebox{\hsize}{!}{ 
\begin{tabular}{lcccccc}
\toprule
Data set & SOG-BC & SOG-GAIL ($\lambda_S=1$)  & InfoGAIL & VAE-GAIL & Expert\\
\midrule
Circles    & \bm{$992.1 \pm 1.0 $} & $985.9\pm10.8$ & $766.0 \pm 67.9$  & $912.3 \pm 8.6$ & $998.3\pm0.1$ \\
Ant-Fwd-Back & \bm{$1165.2 \pm    32.1$} & $1101.0\pm61.7$ &  $220.6 \pm 296.3$ & $-385.3 \pm 67.0$ & $1068.7 \pm 109.7$ \\
Ant-Dir-6    &  \bm{$1073.2 \pm      206.6$} & $1023.2 \pm 166.1$ &    $-14.5 \pm 87.6$    & $-572.9 \pm 62.9$ & $1031.7 \pm 253.7$ \\
HalfCheetah-Fwd-Back & $221.6 \pm 957.3$ & \bm{$1532.6 \pm 148.5$} &  $484.2 \pm 919.4$ & $84.0\pm 152.2$ & $1686.0 \pm 135.4$ \\
Humanoid-Dir-6 & \bm{$5996.0 \pm  326.4$} & $5457.8 \pm 640.9$ &  $ 1333.9 \pm 461.4$ & $2285.5 \pm 946.1$ & $ 6206.6 \pm 292.6$ \\
Walker2d-Vel-6 & \bm{$1915.3 \pm 402.1$} & $1698.7 \pm 650.0$ &  $947.3 \pm 105.3$ & $1183.6 \pm 68.0$ & $1964.6 \pm 394.6$ \\
Hopper-Vel-6 & \bm{$2222.3 \pm 431.1$} & $2015.3 \pm 547.3$ &  $1216.8\pm 70.8$ & $1065.8 \pm 30.1$ & $2229.7 \pm 438.6$ \\
\bottomrule
\end{tabular}
}
\end{center}
\end{table*}
\subsubsection*{Visualization} 
\begin{figure*}[htp]
\centering

\scalebox{0.9}{
\begin{tabular}{@{}l@{}c@{}c@{}c@{}c@{}c@{}}
\rotatebox[origin=c]{90}{\small{Circles}} & 
\raisebox{-0.5\height}{\adjincludegraphics[width=.19\textwidth,trim={{.05\width} {.05\height} {.05\width} {.05\height}},clip]{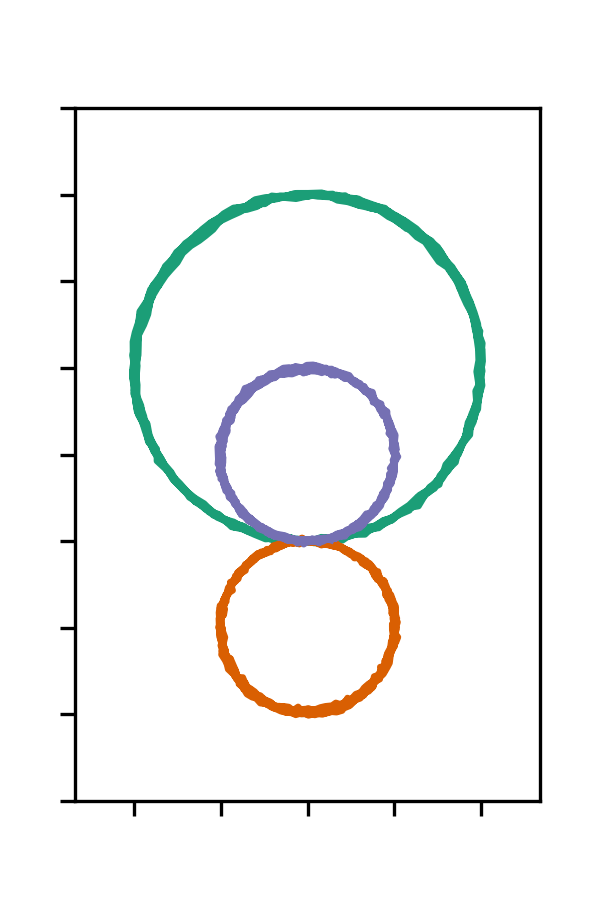}} &
\raisebox{-0.5\height}{\adjincludegraphics[width=.19\textwidth,trim={{.05\width} {.05\height} {.05\width} {.05\height}},clip]{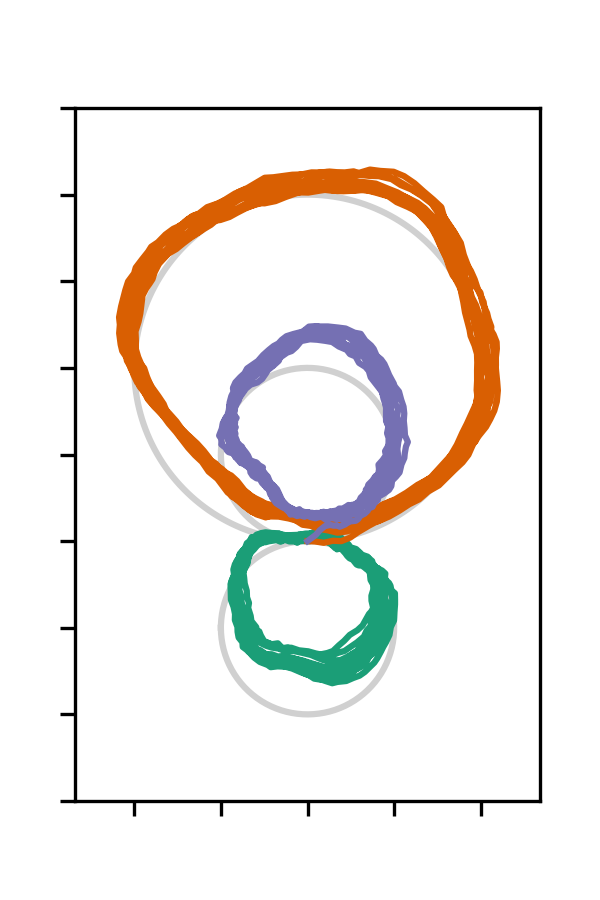}} &
\raisebox{-0.5\height}{\adjincludegraphics[width=.19\textwidth,trim={{.05\width} {.05\height} {.05\width} {.05\height}},clip]{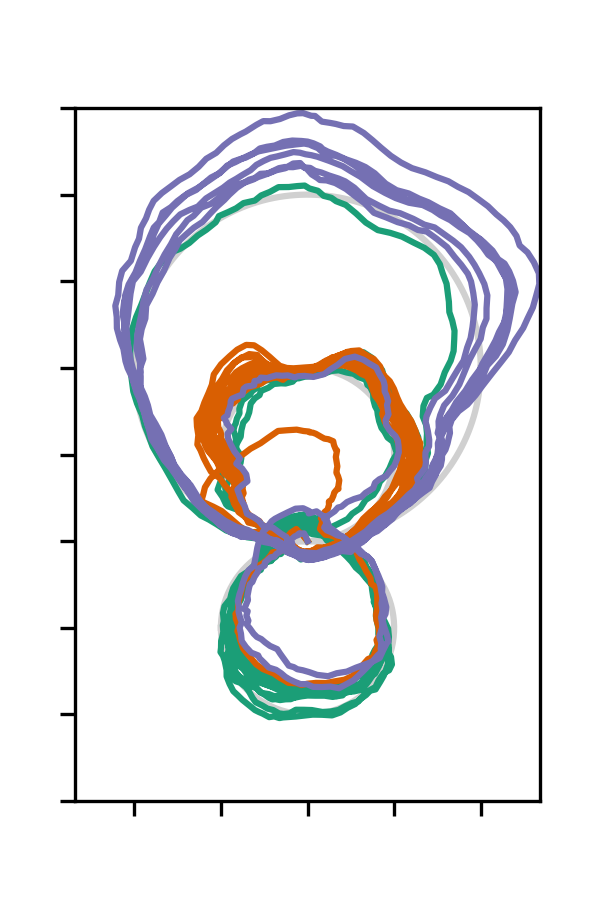}} &
\raisebox{-0.5\height}{\adjincludegraphics[width=.19\textwidth,trim={{.05\width} {.05\height} {.05\width} {.05\height}},clip]{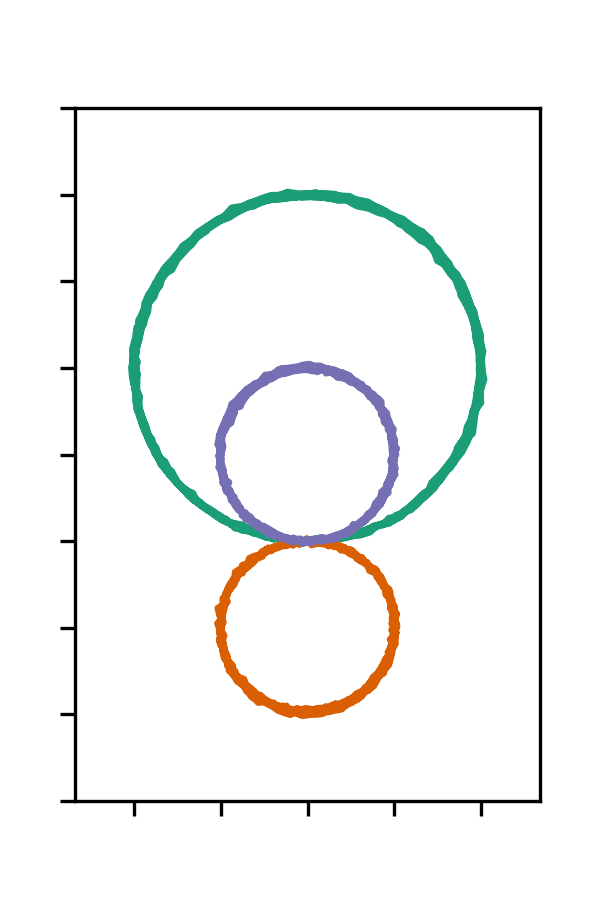}} &
\raisebox{-0.5\height}{\adjincludegraphics[width=.19\textwidth,trim={{.05\width} {.05\height} {.05\width} {.05\height}},clip]{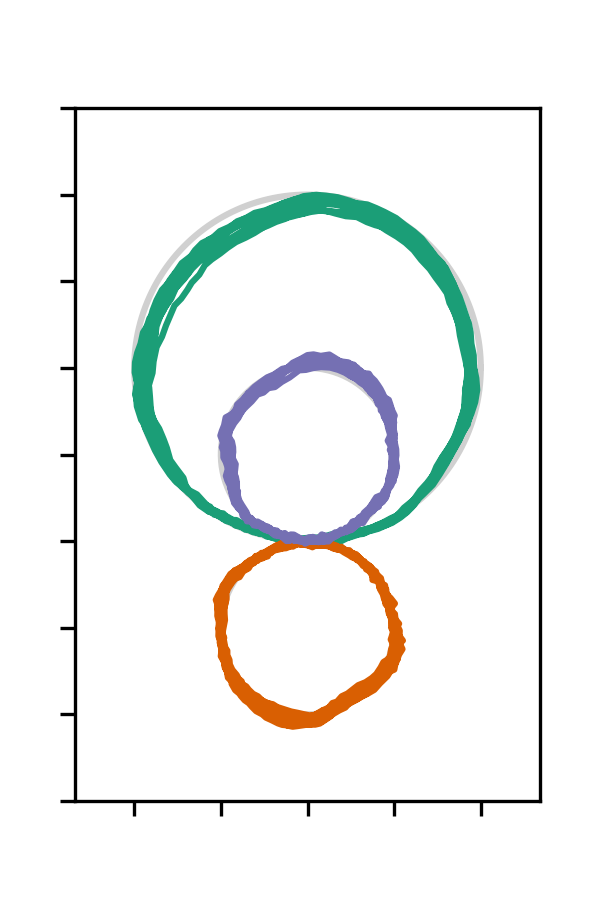}}
\\
\rotatebox[origin=c]{90}{\small{Ant-Fwd-Back}} &
\raisebox{-0.5\height}{\adjincludegraphics[width=.19\textwidth,trim={{.05\width} {.05\height} {.05\width} {.05\height}},clip]{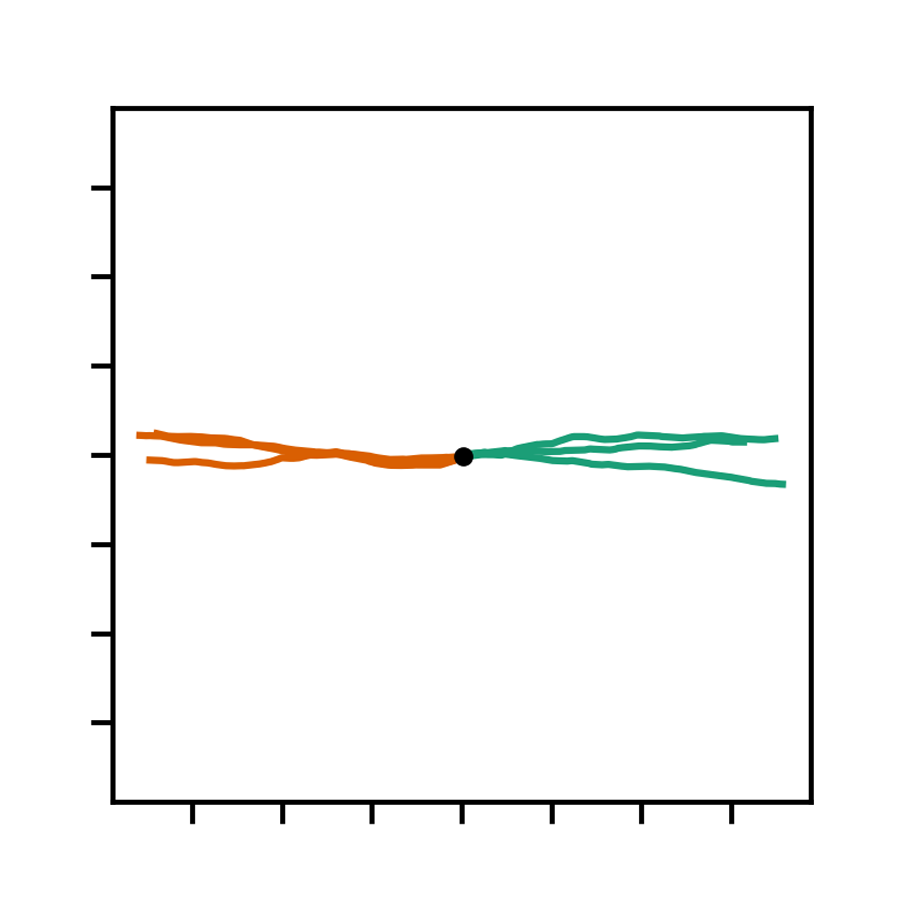}} &
\raisebox{-0.5\height}{\adjincludegraphics[width=.19\textwidth,trim={{.05\width} {.05\height} {.05\width} {.05\height}},clip]{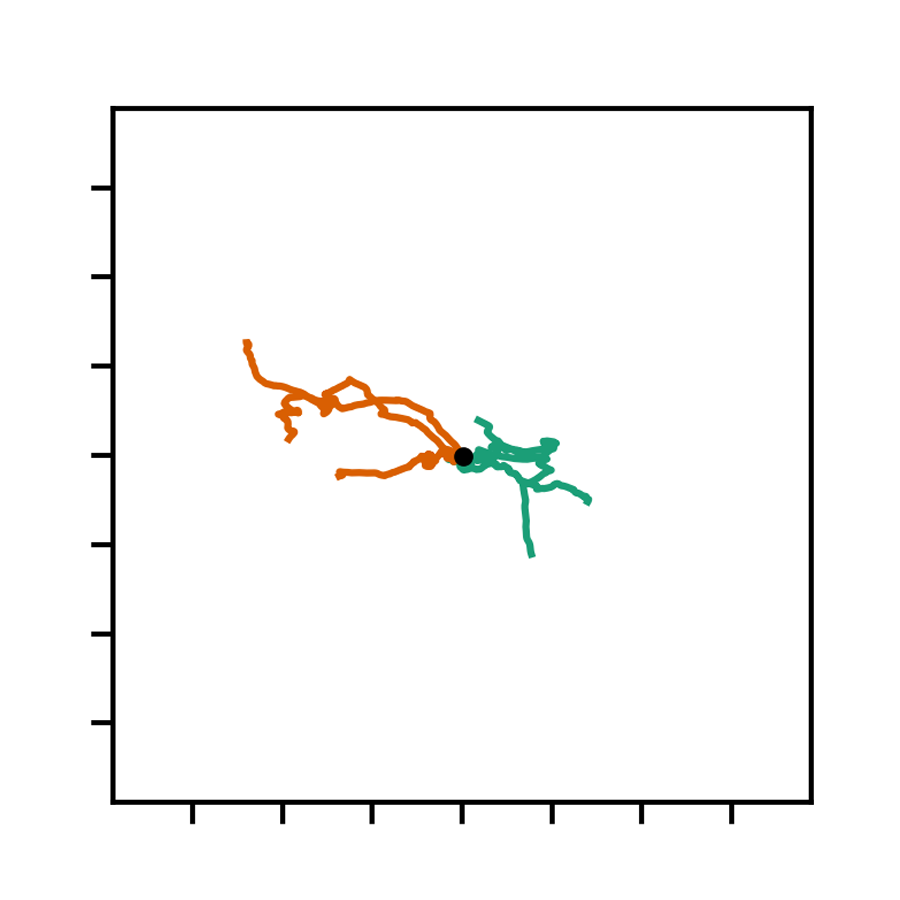}} &
\raisebox{-0.5\height}{\adjincludegraphics[width=.19\textwidth,trim={{.05\width} {.05\height} {.05\width} {.05\height}},clip]{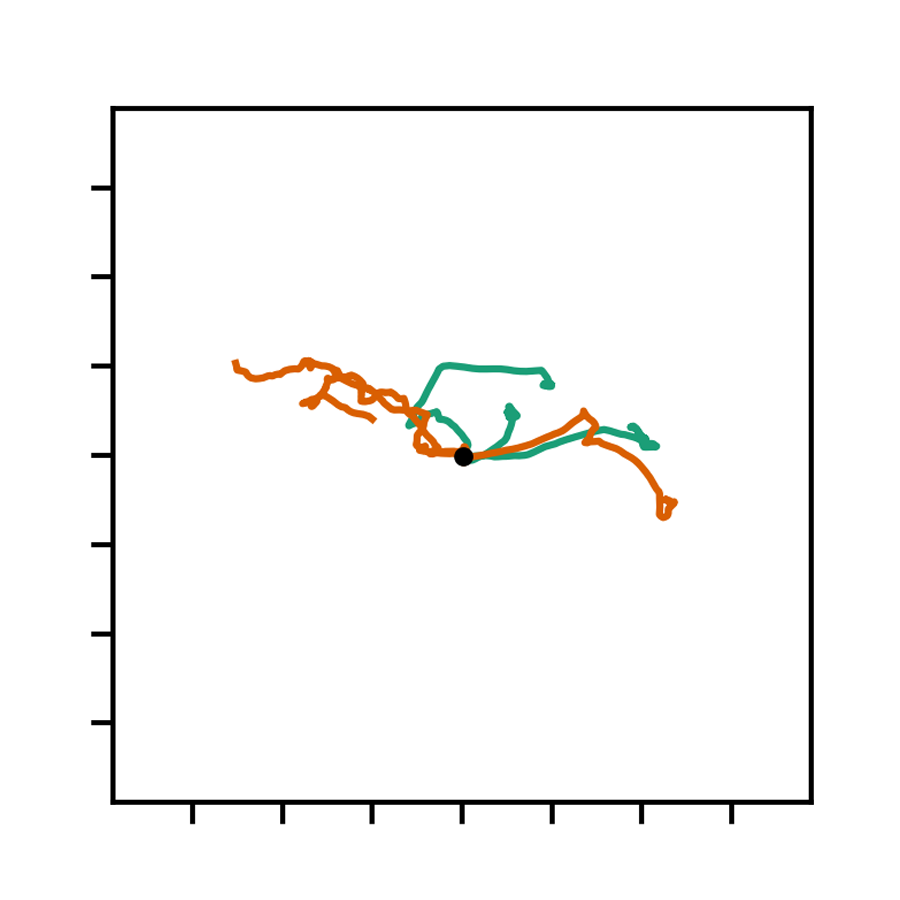}} &
\raisebox{-0.5\height}{\adjincludegraphics[width=.19\textwidth,trim={{.05\width} {.05\height} {.05\width} {.05\height}},clip]{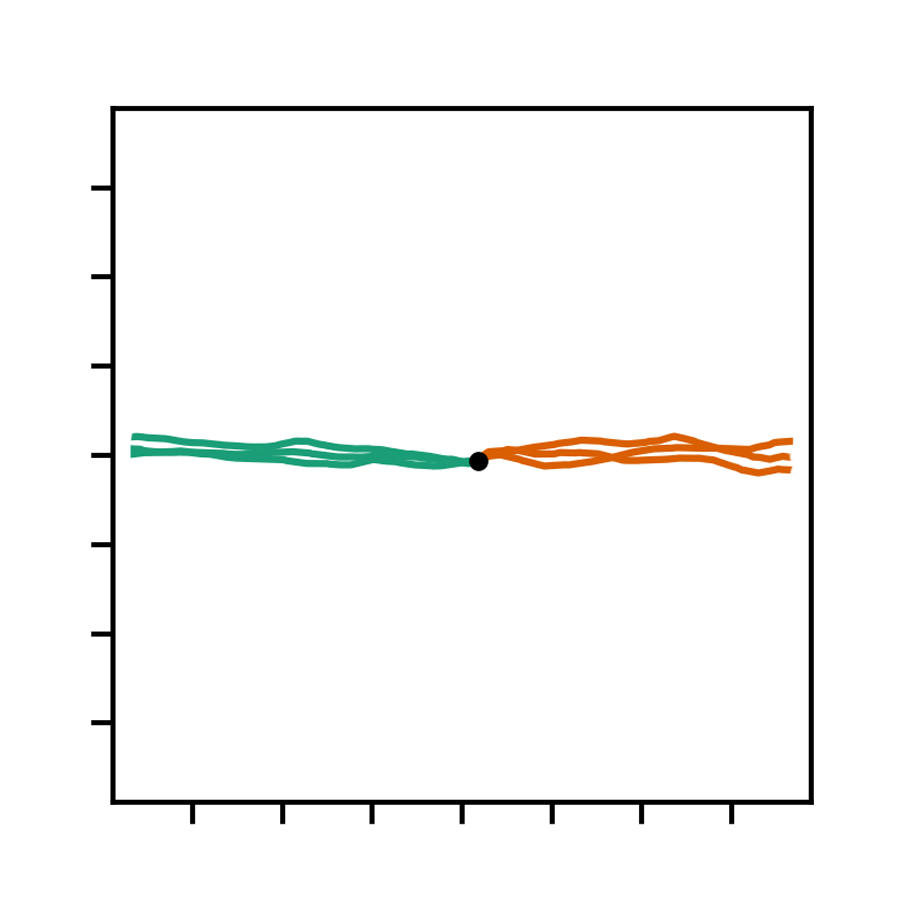}} &
\raisebox{-0.5\height}{\adjincludegraphics[width=.19\textwidth,trim={{.05\width} {.05\height} {.05\width} {.05\height}},clip]{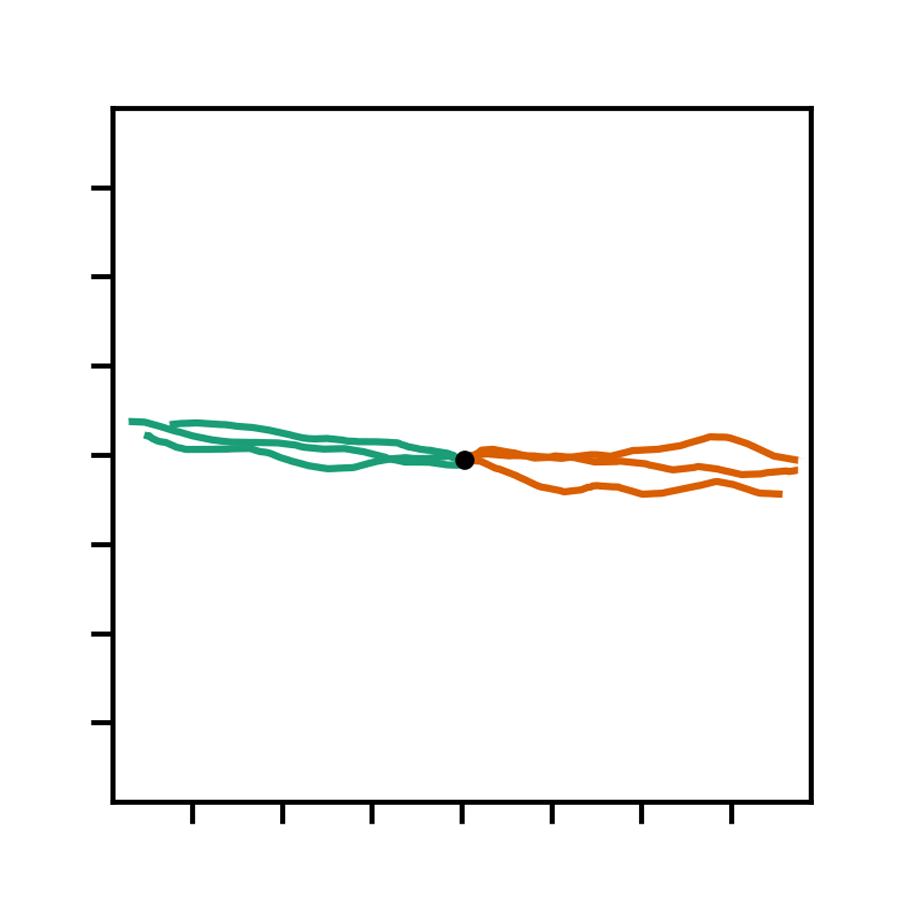}} 
\\
\rotatebox[origin=c]{90}{\small{Ant-Dir-6}} & 
\raisebox{-0.5\height}{\adjincludegraphics[width=.19\textwidth,trim={{.05\width} {.05\height} {.05\width} {.05\height}},clip]{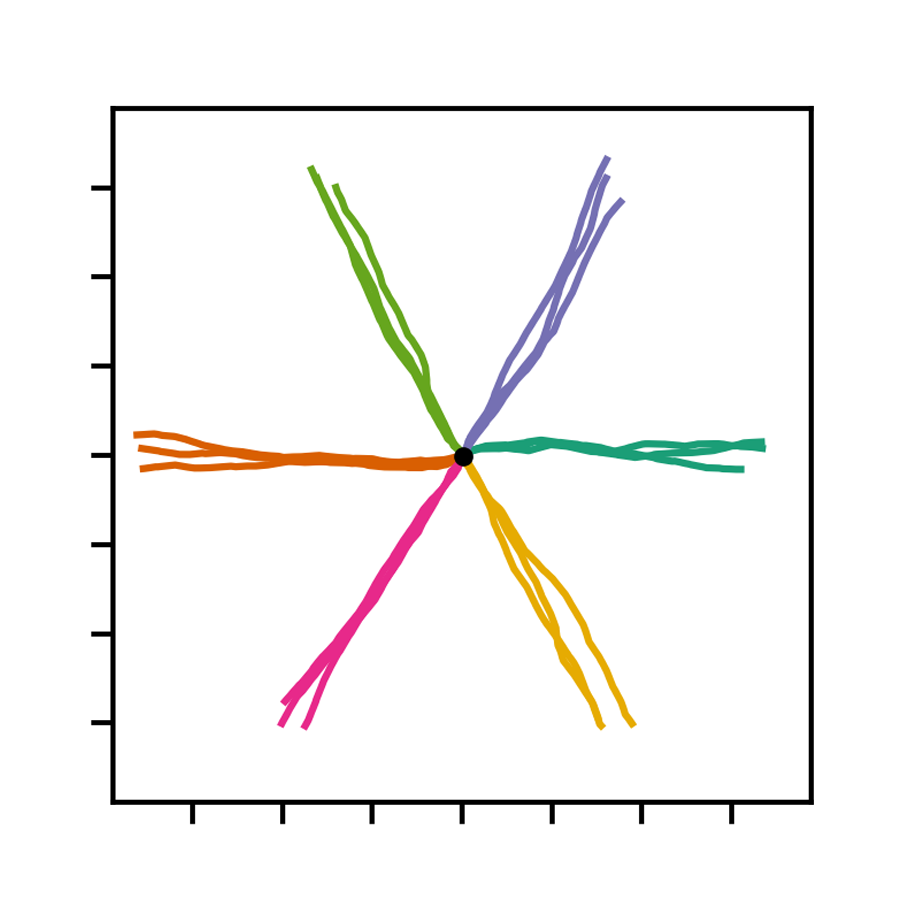}} & 
\raisebox{-0.5\height}{\adjincludegraphics[width=.19\textwidth,trim={{.05\width} {.05\height} {.05\width} {.05\height}},clip]{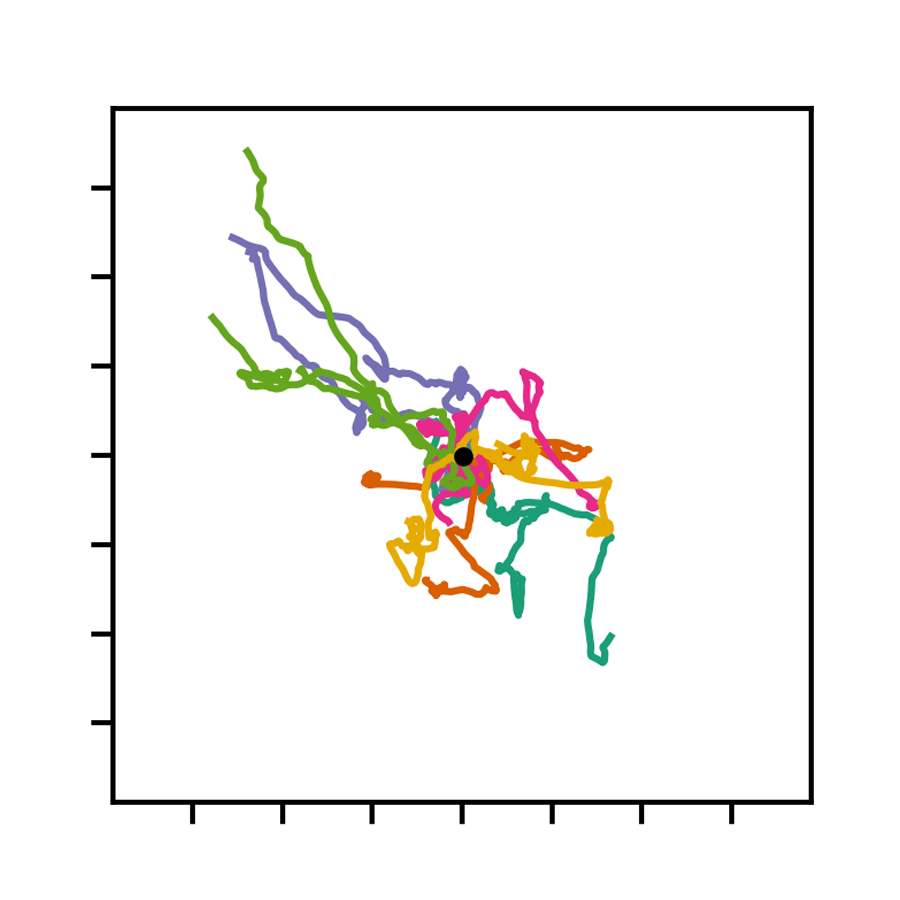}} &
\raisebox{-0.5\height}{\adjincludegraphics[width=.19\textwidth,trim={{.05\width} {.05\height} {.05\width} {.05\height}},clip]{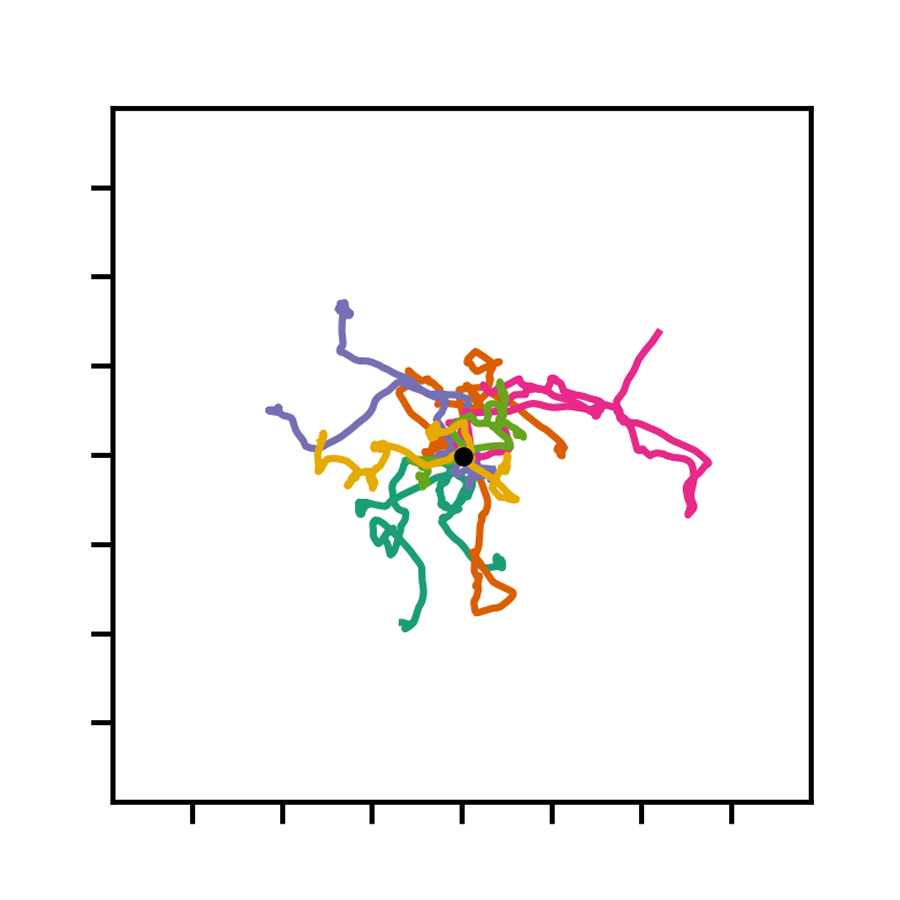}} &
\raisebox{-0.5\height}{\adjincludegraphics[width=.19\textwidth,trim={{.05\width} {.05\height} {.05\width} {.05\height}},clip]{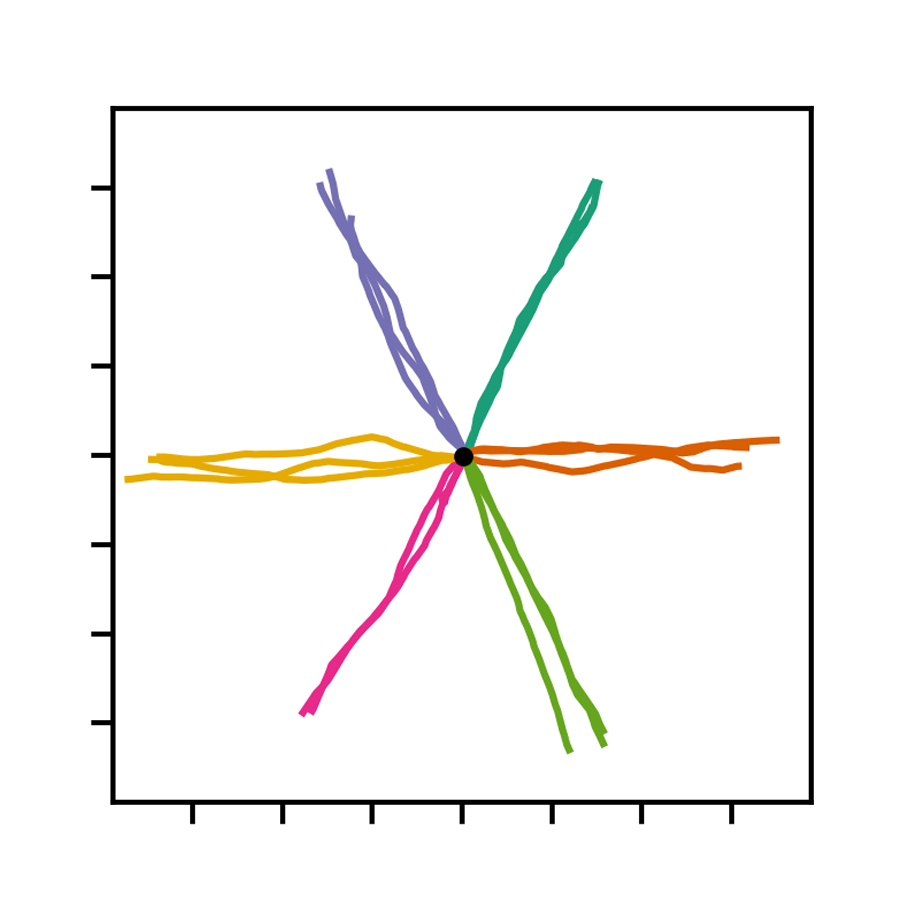}} &
\raisebox{-0.5\height}{\adjincludegraphics[width=.19\textwidth,trim={{.05\width} {.05\height} {.05\width} {.05\height}},clip]{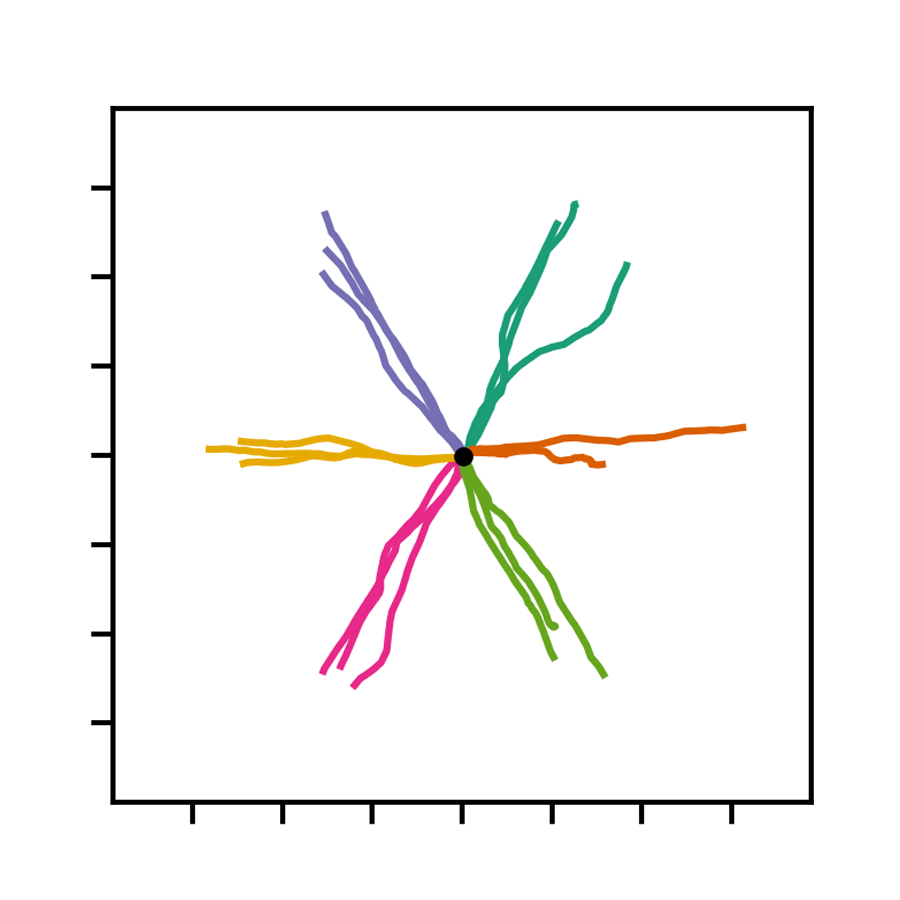}} 
\\
\rotatebox[origin=c]{90}{\small{Humanoid-Dir-6}} & 
\raisebox{-0.5\height}{\adjincludegraphics[width=.19\textwidth,trim={{.05\width} {.05\height} {.05\width} {.05\height}},clip]{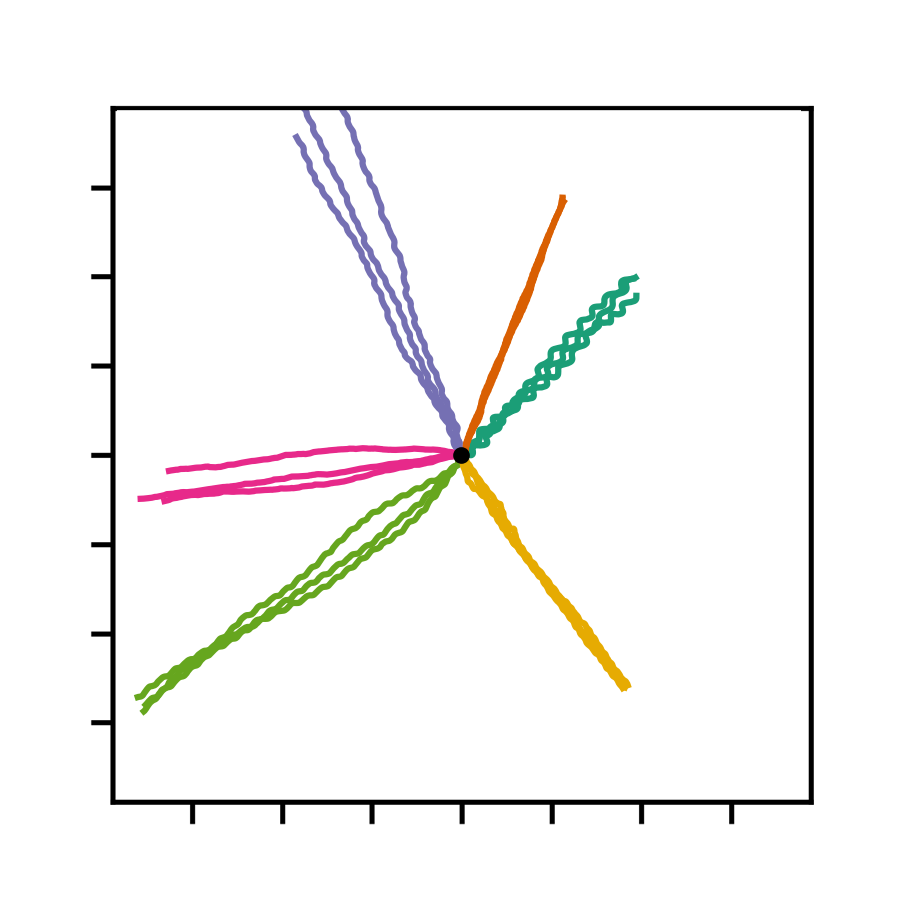}} &
\raisebox{-0.5\height}{\adjincludegraphics[width=.19\textwidth,trim={{.05\width} {.05\height} {.05\width} {.05\height}},clip]{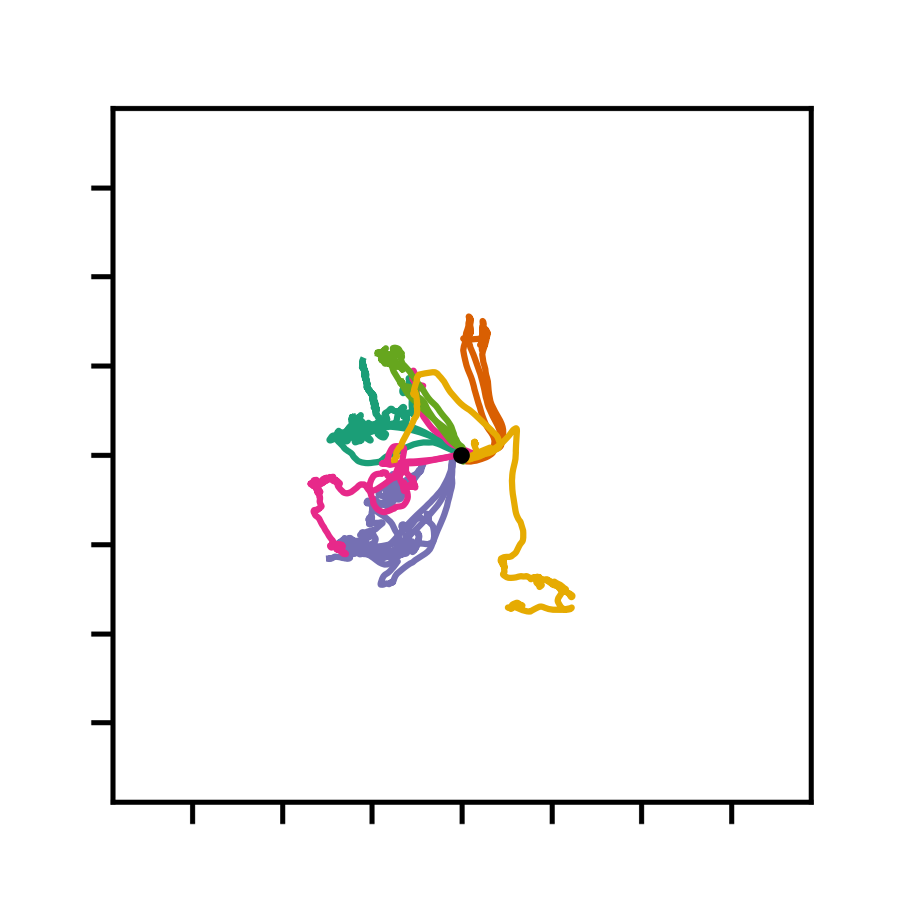}} &
\raisebox{-0.5\height}{\adjincludegraphics[width=.19\textwidth,trim={{.05\width} {.05\height} {.05\width} {.05\height}},clip]{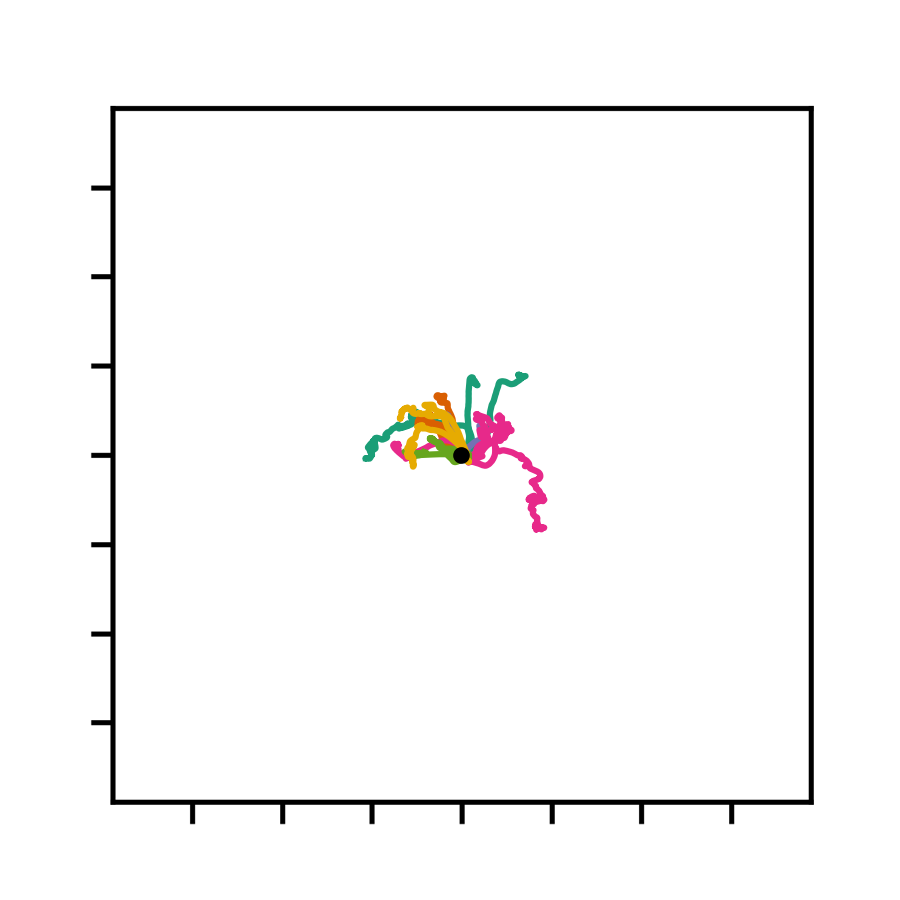}} &
\raisebox{-0.5\height}{\adjincludegraphics[width=.19\textwidth,trim={{.05\width} {.05\height} {.05\width} {.05\height}},clip]{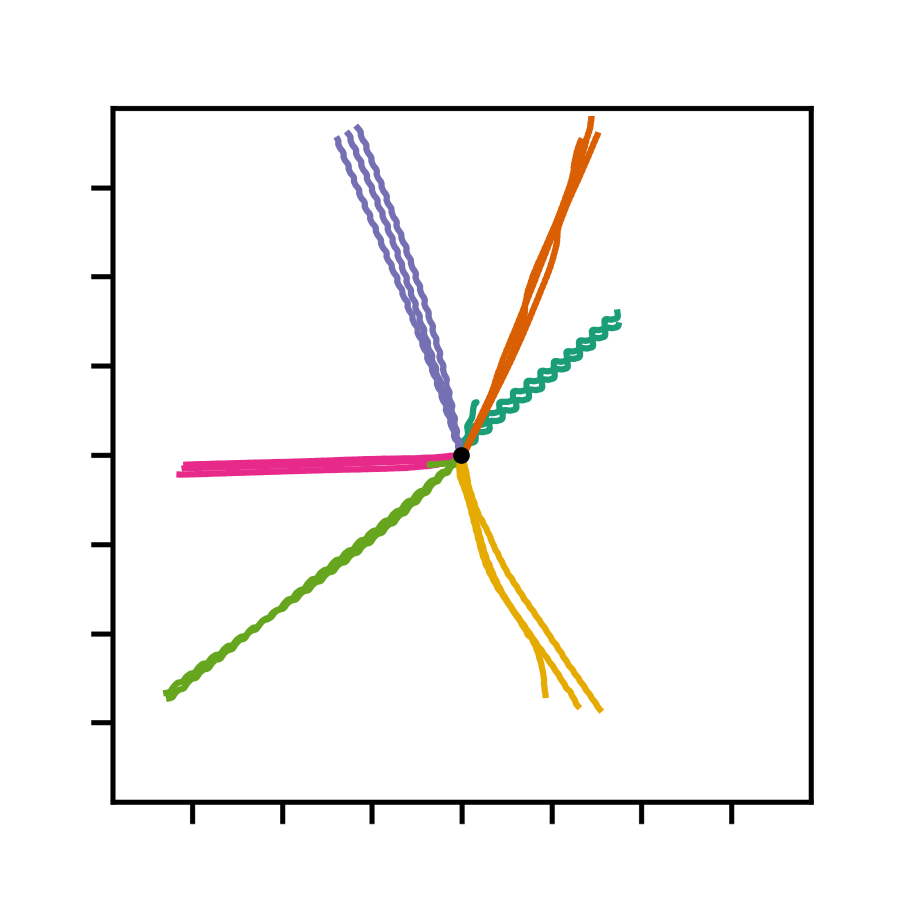}} &
\raisebox{-0.5\height}{\adjincludegraphics[width=.19\textwidth,trim={{.05\width} {.05\height} {.05\width} {.05\height}},clip]{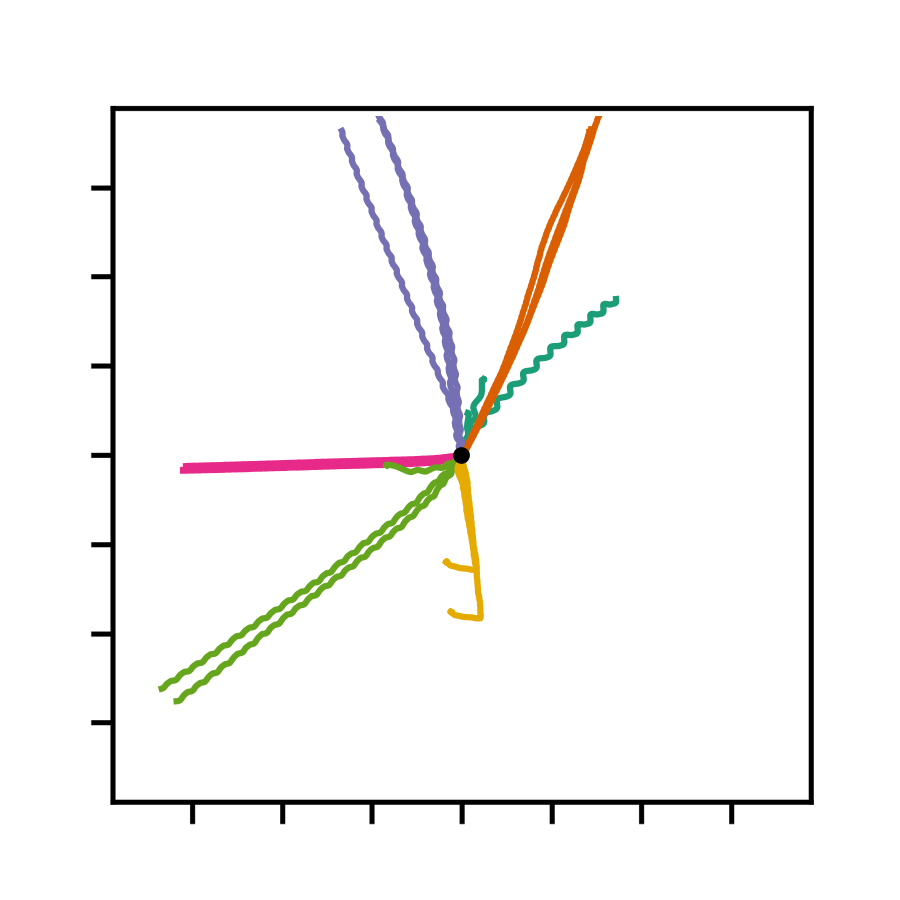}} 
\\
& \small{Expert}
& \small{VAE-GAIL}
& \small{InfoGAIL}
& \small{SOG-BC}
& \small{SOG-GAIL ($\lambda_S=1$)}
\end{tabular}
}
\caption{\label{fig:trajs} \small  \textbf{Discrete latent variable: locomotion towards multiple directions.} Trajectories of the imitated policies for four tasks. Different latent codes at inference time are color coded. From top to bottom: number of modes are 3, 2, 6, 6; number of trajectories per latent code are 1, 3, 3, 3. Both SOG-BC and SOG-GAIL precisely separate and imitate the modes in all the experiments. However, VAE-GAIL and InfoGAIL fail to separate the modes and to imitate the expert.}
\vspace{-10pt}
\end{figure*}

\begin{figure*}[t]
\centering
\begin{tabular}{@{}l@{}c@{}c@{}c@{}c@{}c@{}}
\rotatebox[origin=c]{90}{\small{Walker2d-Vel-6}} & 
\raisebox{-0.5\height}{\adjincludegraphics[width=.19\textwidth,trim={{.01\width} {.0\height} {.08\width} {.1\height}},clip]{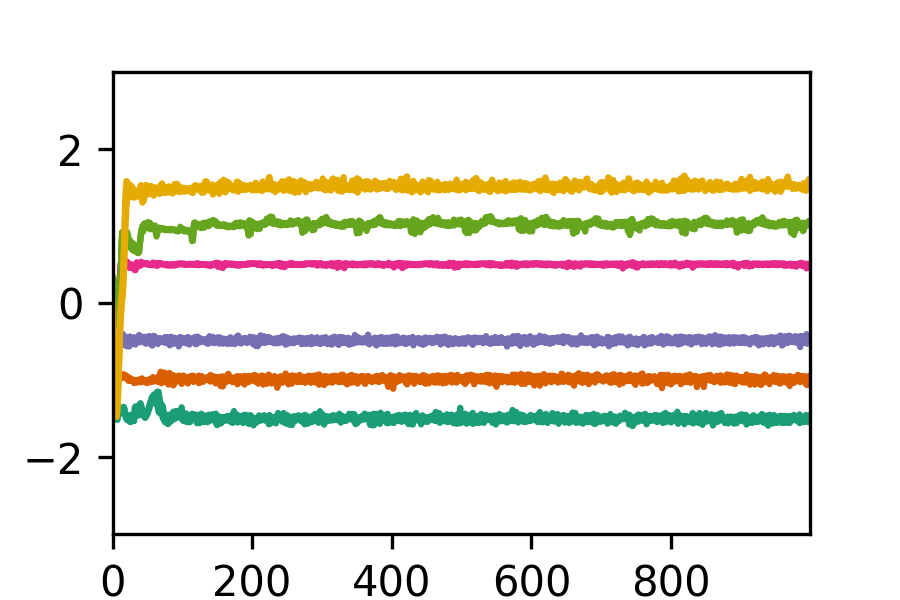}} &
\raisebox{-0.5\height}{\adjincludegraphics[width=.19\textwidth,trim={{.01\width} {.0\height} {.08\width} {.1\height}},clip]{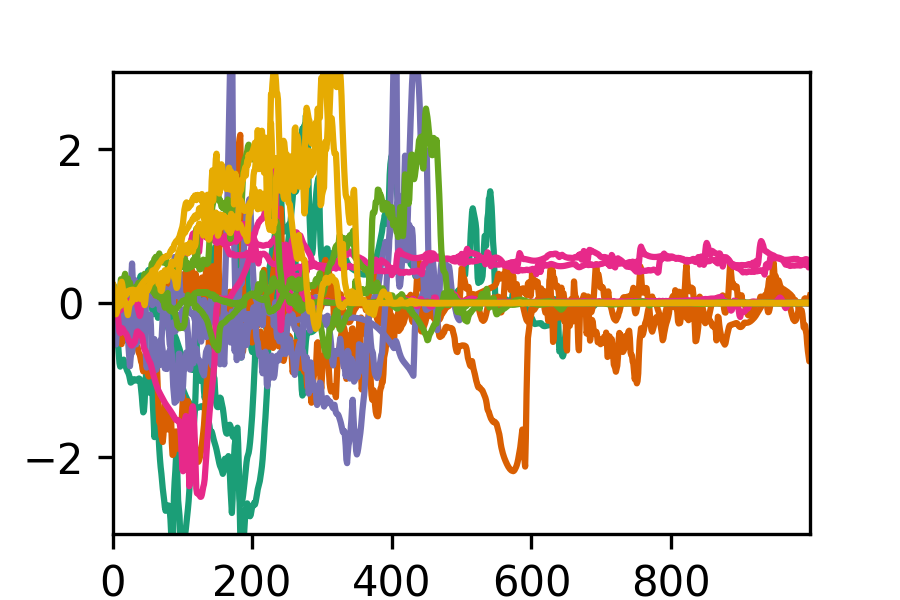}} &
\raisebox{-0.5\height}{\adjincludegraphics[width=.19\textwidth,trim={{.01\width} {.0\height} {.08\width} {.1\height}},clip]{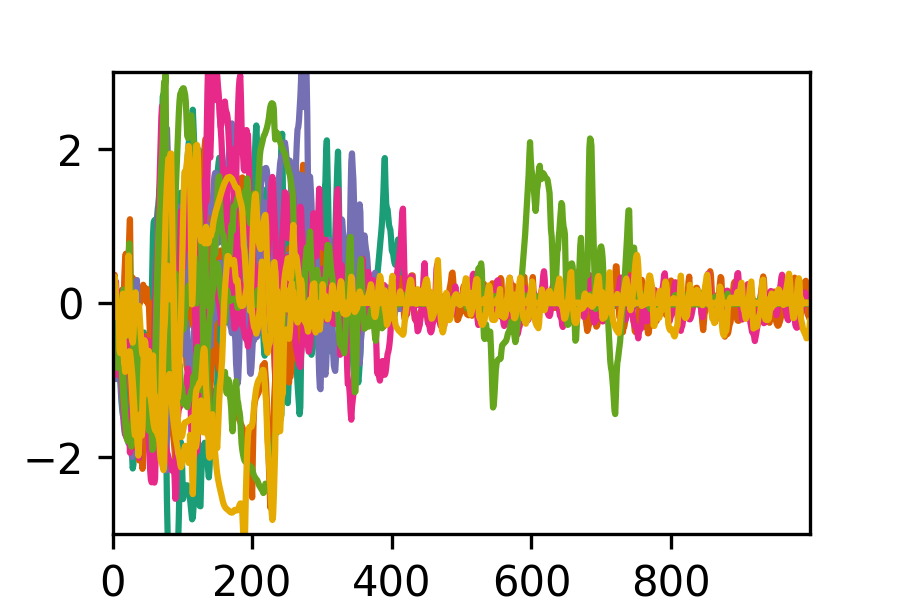}} &
\raisebox{-0.5\height}{\adjincludegraphics[width=.19\textwidth,trim={{.01\width} {.0\height} {.08\width} {.1\height}},clip]{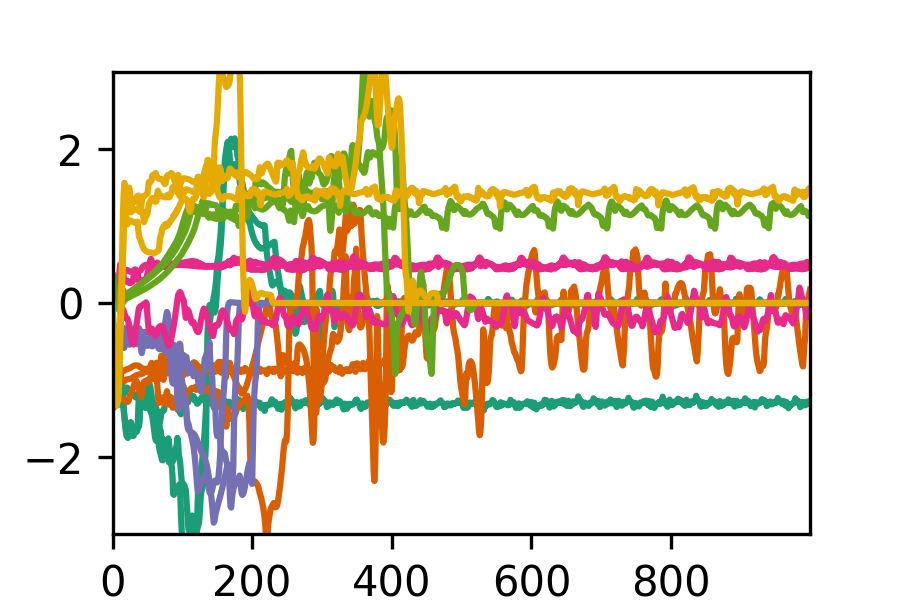}} &
\raisebox{-0.5\height}{\adjincludegraphics[width=.19\textwidth,trim={{.01\width} {.0\height} {.08\width} {.1\height}},clip]{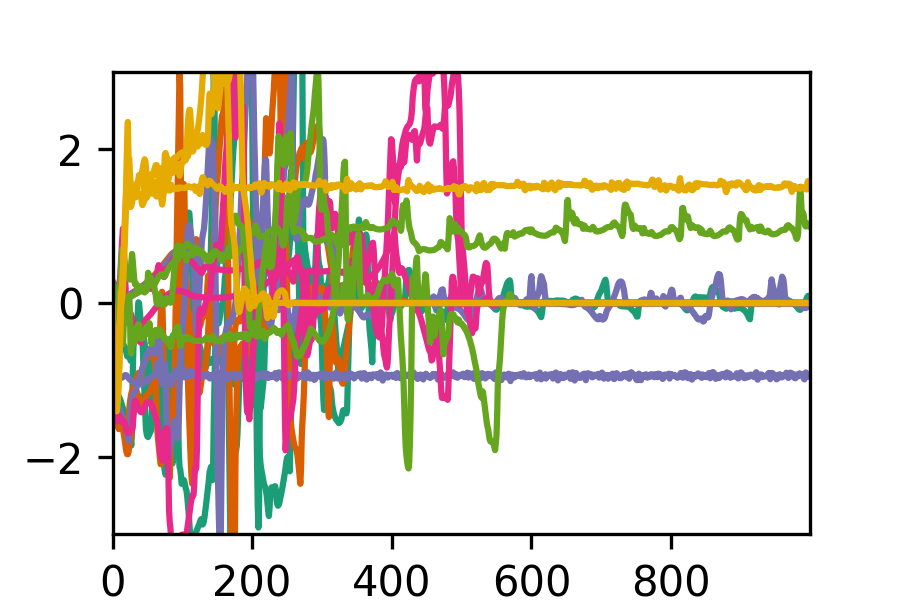}}
\\
\addlinespace[0.2cm]
\rotatebox[origin=c]{90}{\small{Hopper-Vel-6}} &
\raisebox{-0.5\height}{\adjincludegraphics[width=.19\textwidth,trim={{.01\width} {.0\height} {.08\width} {.1\height}},clip]{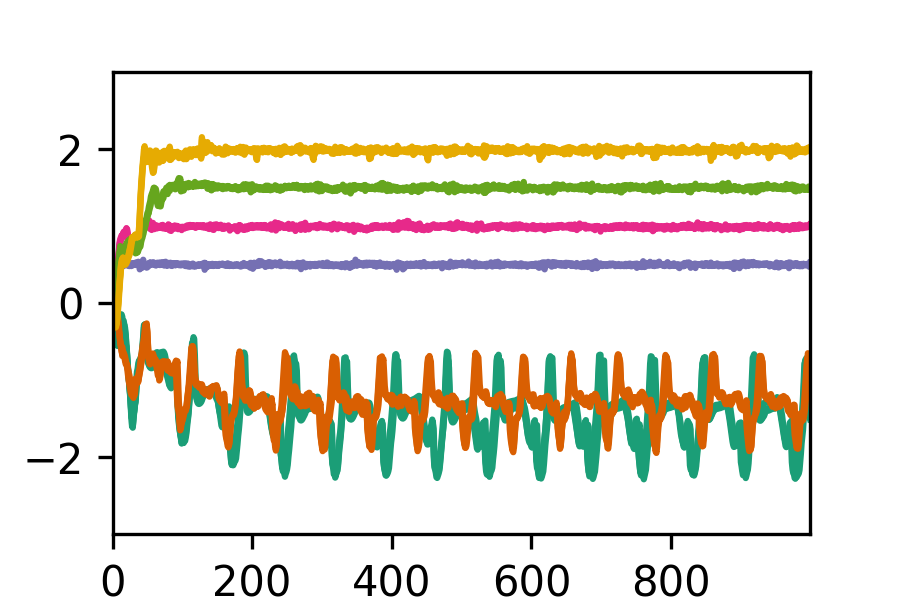}} &
\raisebox{-0.5\height}{\adjincludegraphics[width=.19\textwidth,trim={{.01\width} {.0\height} {.08\width} {.1\height}},clip]{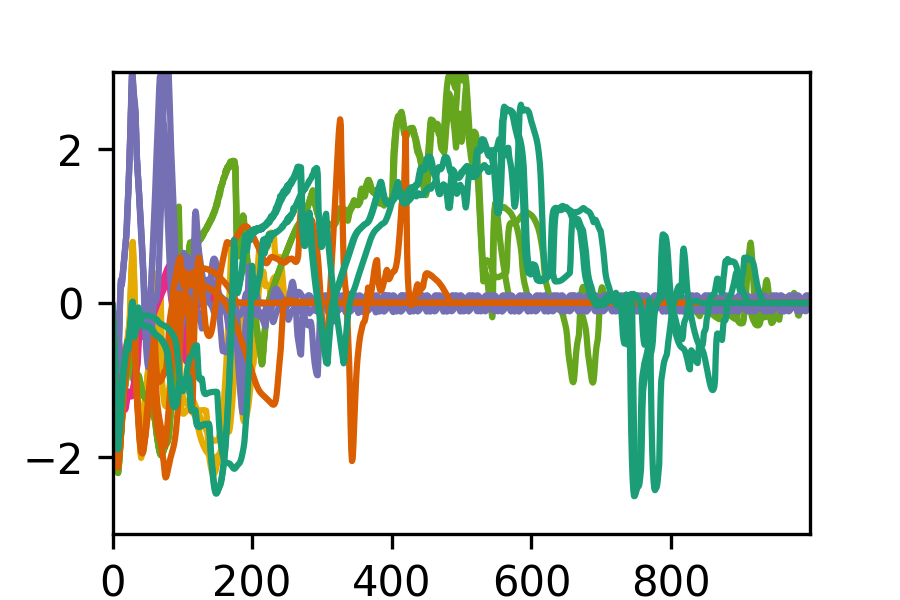}} &
\raisebox{-0.5\height}{\adjincludegraphics[width=.19\textwidth,trim={{.01\width} {.0\height} {.08\width} {.1\height}},clip]{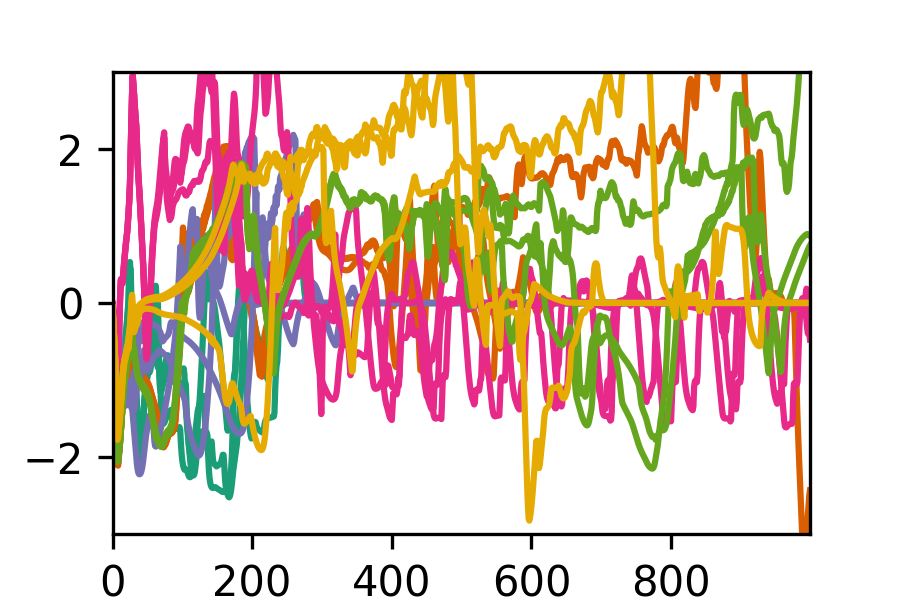}} &
\raisebox{-0.5\height}{\adjincludegraphics[width=.19\textwidth,trim={{.01\width} {.0\height} {.08\width} {.1\height}},clip]{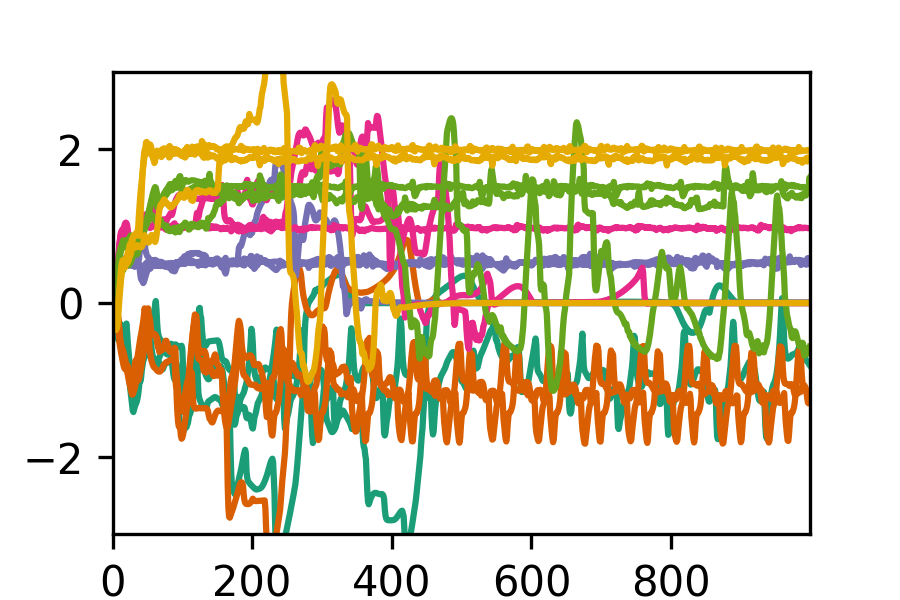}} &
\raisebox{-0.5\height}{\adjincludegraphics[width=.19\textwidth,trim={{.01\width} {.0\height} {.08\width} {.1\height}},clip]{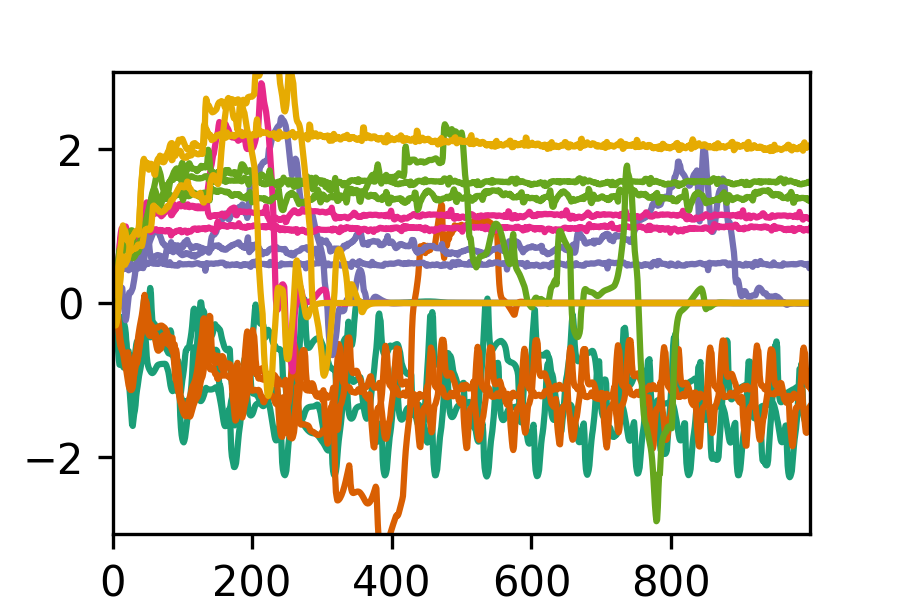}} 
\\
& \small{Expert}
& \small{VAE-GAIL}
& \small{InfoGAIL}
& \small{SOG-BC}
& \small{SOG-GAIL ($\lambda_S=1$)}
\end{tabular}
\caption{\label{fig:vels} \small  \textbf{Discrete latent variable: locomotion at six velocities.} 
Velocities generated by the policies vs rollout time steps. Three rollouts per mode are visualized. Each mode is marked with a distinct color. Both SOG-BC and SOG-GAIL separate and imitate most of the modes, producing the desired velocities.  VAE-GAIL and InfoGAIL exhibit an inferior performance.
}
\vspace{-10pt}
\vspace{1em}
\input{results/fetch}
\end{figure*}

In \Cref{fig:trajs}, we visualize the generated trajectories of four locomotion tasks with discrete modes.
Throughout this figure, we can see that both SOG-BC and SOG-GAIL successfully learn the different modes. Moreover, we observe that SOG-BC performs slightly more accurately than  SOG-GAIL. We discuss the differences between the two algorithms in a later paragraph, where we compare them in terms of their robustness. In \Cref{fig:vels}, we visualize the velocities of Walker2d-Vel-6 and Hopper-Vel-6 experiments under different policies. We observe that while SOG-BC successfully distinguishes and reconstructs the desired velocities, the baseline models fail.
\begin{table}
\caption{Metrics for reached targets in the FetchReach experiment: average hit rate and estimated entropy of achieved targets (higher is better).
}
\label{tab:entropy}
\begin{center}
\begin{scriptsize}
\resizebox{\columnwidth}{!}{ 
\begin{tabular}{lccccc}
\toprule
Metric & SOG-BC & SOG-GAIL  & InfoGAIL & VAE-GAIL & Expert\\
\midrule
Entropy (nats) &\bm{$2.05$} & $1.89$ & $0.27$  & $0.85$ & $2.18$ \\
Hit Rate & \bm{$100\%$} & $97.0\%$ & N/A  & $18.6\%$ & $100\%$ \\
\bottomrule
\end{tabular}
}
\end{scriptsize}
\end{center}
\end{table}

\begin{table}
\caption{Mutual information between the latent variable (target velocity of the embedded trajectory in the case of VAE-GAIL) and generated velocities in HalfCheetah-Vel.}
\label{tab:mutual-info}
\begin{center}
\begin{tabular}{ccccc}
\toprule
SOG-BC & SOG-GAIL  & InfoGAIL & VAE-GAIL\\
\midrule
\bm{$1.584$} & $1.431$ & $0.145$  & $0.750$ \\
\bottomrule
\end{tabular}
\end{center}
\end{table}

\begin{figure*}
     \centering
     \begin{subfigure}[b]{0.28\textwidth}
         \centering
         \adjincludegraphics[width=\textwidth,trim={{.00\width} {0\height} {.05\width} {.05\height}},clip]{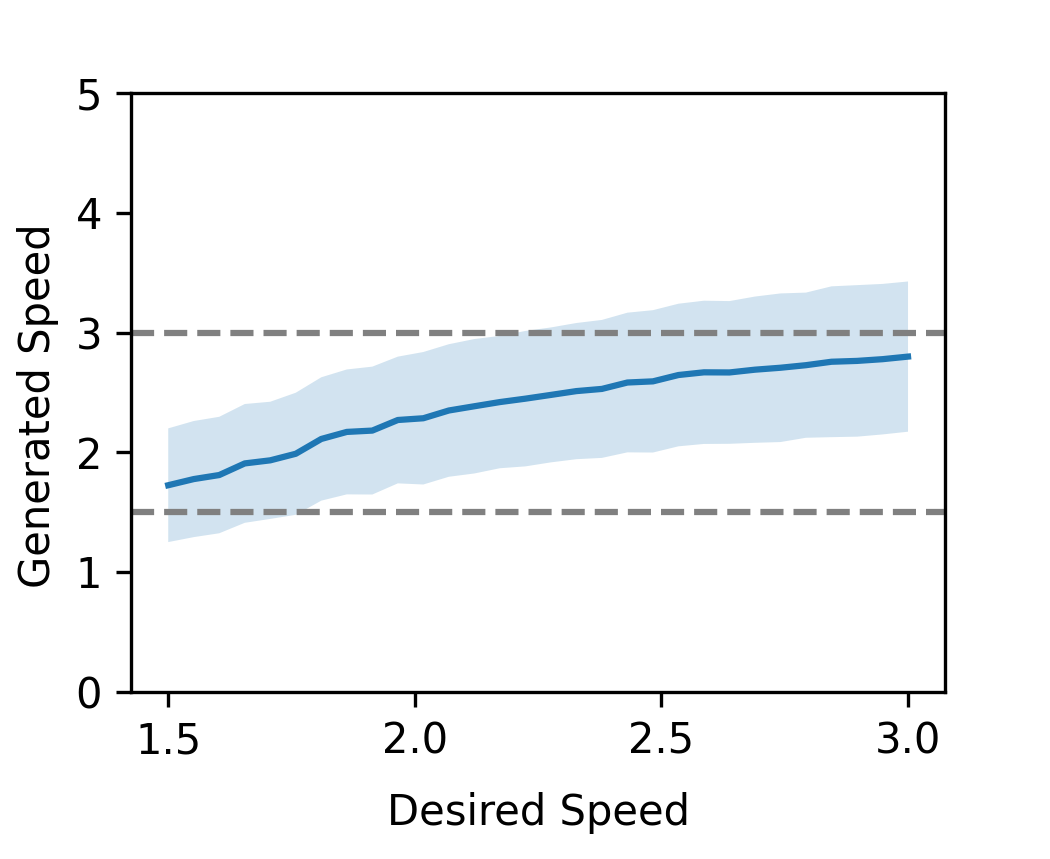}
         \caption{\label{fig:hcv-VAE-GAIL}VAE-GAIL}
     \end{subfigure}
     \hfill
         \begin{subfigure}[b]{0.28\textwidth}
         \centering
         \adjincludegraphics[width=\textwidth,trim={{.00\width} {0\height} {.05\width} {.05\height}},clip]{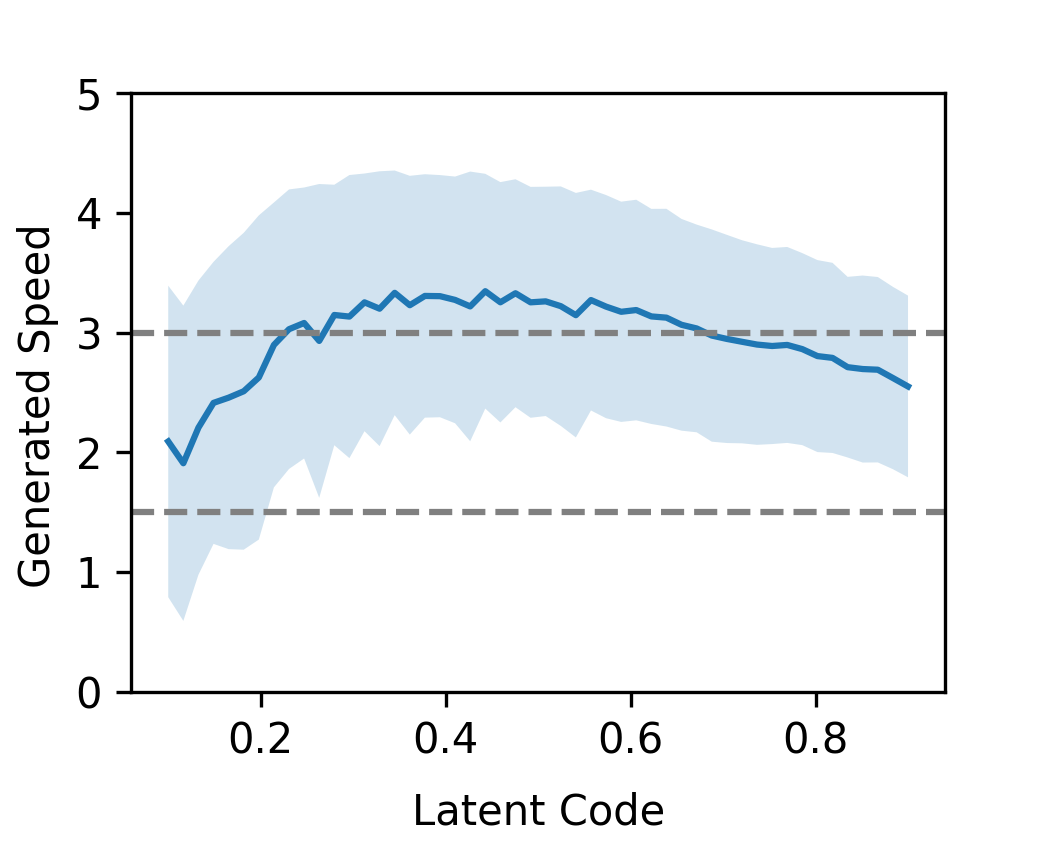}
         \caption{InfoGAIL}
     \end{subfigure}
     \hfill
     \begin{subfigure}[b]{0.28\textwidth}
         \centering
         \adjincludegraphics[width=\textwidth,trim={{.00\width} {0\height} {.05\width} {.05\height}},clip]{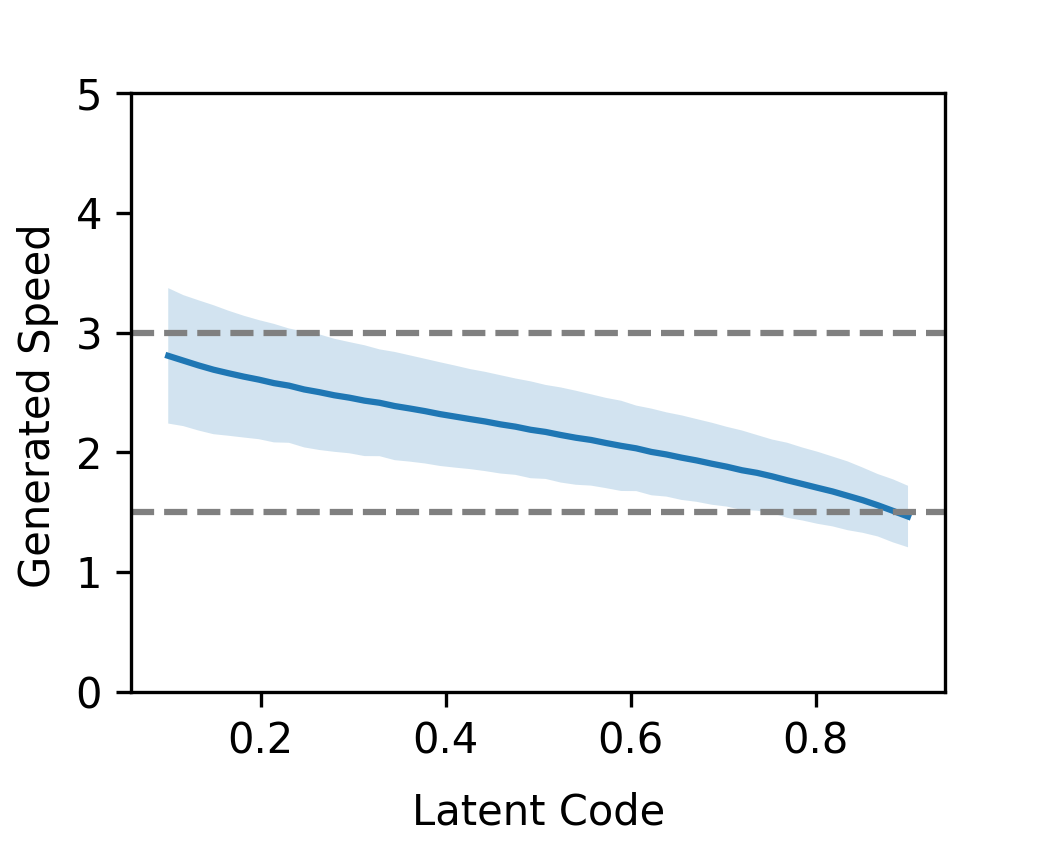}
         \caption{\label{fig:hcv-SOG}SOG-BC}
     \end{subfigure}

        \caption{ \small \textbf{Continuous latent variable: HalfCheetah-Vel.}
        In (a), since VAE-GAIL uses a high latent dimension, in the horizontal axis we consider the desired velocity of the expert trajectory corresponding to the latent embedding. However, in (b), (c) the horizontal axis corresponds to (the CDF of) the 1-D Gaussian latent variable. The CDF is applied for better visualization. In SOG-BC, different velocities in range $[1.5,3]$ are replicated with a higher certainty. 
        }
\label{fig:hcv} 
\end{figure*}

In \Cref{fig:fetch,fig:fetch-latent,fig:hcv}, we visualize the results of continuous experiments FetchReach and \hbox{HalfCheetah-Vel}. In these figures, SOG-BC and SOG-GAIL produce similar plots, hence we drop the SOG-GAIL plot.

In \Cref{fig:fetch}, we plot the trajectories of the Fetch robotic arm for different samples of the latent variable. We observe that SOG-BC is able to control the robotic arm towards any point in the 3D space. In \Cref{fig:fetch-latent}, we plot the reached targets for different choices of the latent code, and observe a semantically meaningful interpolation in the latent space. This \href{https://youtu.be/ecI9Phj5N0w}{\textbf{video}} contains animations of test rollouts.

In \Cref{fig:hcv}, we plot the spread of instantaneous horizontal velocities of the HalfCheetah agent as the latent variable varies. In SOG-BC and InfoGAIL, the horizontal axis corresponds to the 1-D latent code used to train the model. However, since VAE-GAIL requires a high dimensional latent variable for best performance, the correspondence between latent codes and the model generated horizontal velocities cannot be directly visualized. Therefore, we utilize the following fact: if the latent embedding of an expert trajectories is fed to the learned policy, the generated trajectory produces similar velocities as the expert trajectory. Thus, for VAE-GAIL we plot the variations in the generated velocities as a function of the target velocity of the corresponding expert trajectory. We observe that compared to the baselines, SOG-BC better associates the latent variable to goal velocities in the range $[1.5,3]$. Also, SOG-BC produces a narrower error band, which indicates better certainty given fixed latent codes.  This \href{https://youtube.com/shorts/SBW2Nu-fisg}{\textbf{video}} illustrates the imitated behavior of SOG-BC at different velocities as the latent code varies.


\subsubsection*{Metrics} In \Cref{tab:rewards}, we use ground-truth reward functions to evaluate the collected rewards under different policies. We explain these ground-truth rewards in \Cref{appx:experiment-details}. In each entry of the table, we report the best iteration in terms of the mean reward. Particularly, we calculate mean and std of the rewards across 100 rollouts. To find the mode-specific reward for each latent code, we select a correspondence of latent codes to actual modes that gives the best expected sum of rewards. This metric measures whether all modes are separated and properly imitated, because each mode has a distinct reward function. Across all experiments, SOG-BC performs significantly better than the baselines. The only exception is HalfCheetah-Dir, where SOG-GAIL performs better. We attribute this latter observation to the difficulty of the task, which is alleviated by the explorations of GAIL.

In \Cref{tab:entropy}, we evaluate the FetchReach experiment through two metrics. First, we collect the achieved targets under different policies, and clip them within the cubic region from which the targets are sampled. Then, we quantify the desired uniformity of the distribution of the achieved targets over the cube. Since uniform distribution is the maximum entropy distribution over rectangular supports, we alternatively  estimate the entropy of the achieved targets via the method of \citet{kozachenko1987sample}. As a second metric, we embed expert trajectories, and measure the hit rate of the reconstructed trajectories given the same targets as the expert.

In the HalfCheetah-Vel experiment, to measure the correspondence between the latent variable and generated instantaneous horizontal velocities, we estimate their mutual information through the method of \citet{kraskov2004estimating}. The results are presented in \Cref{tab:mutual-info}.

\subsubsection*{Robustness}
We design additional experiments to show that integration of SOG with GAIL in \Cref{alg:sog-gail} makes the learned policies robust towards unseen states. Observing space limits, we present the results in \Cref{appx:robust}.




\section{Concluding Remarks} Our empirical results show that SOG-BC and SOG-GAIL distinguish and learn different modes of behavior better than the baseline models, particularly in the discrete case. This advantage can be attributed to the differences between the methods in inducing mode separation in GAIL. Specifically, VAE-GAIL and InfoGAIL directly modify the GAIL optimization to encourage multimodality. However, our method learns multiple modes in GAIL indirectly and through integration with multimodal BC. 

\bibliographystyle{named}
\bibliography{ijcai22}

\clearpage
\appendix
\appendix 
\section{Imitation Learning Algorithms} \label{appx:algorithms}
In this section, we present the imitation learning algorithms described in \Cref{sec:sog-imitation}.
\begin{algorithm}
	\caption{SOG-BC: Multimodal Behavior Cloning with SOG}
	\label{alg:sog-bc}
	\begin{algorithmic}[1]
	    \State \textbf{Define}: Loss function $\mathcal{L}\left(\hat{\va},\va\right)\coloneqq ||\hat{\va}-\va||^2$
	    \State \textbf{Input:} Initial parameters of policy network $\boldsymbol\theta_0$; expert trajectories $\tau_E\sim\pi_E$
		\For {$iteration=0,1,2,\ldots$}
    	    \parState{Sample state-action pairs $\chi_E\sim\tau_E$ with the same batch size.}
	        \parState{Sample $N_z$ latent codes $\displaystyle \vz_i \sim p(\vz)$ for each trajectory.} \label{alg:sog-bc:latent-batch}
            \parState{Calculate $\vz_E\coloneqq\arg\min_{\vz_i}\hat\E_{\tau_E}\left[\mathcal{L}\left(f_{\boldsymbol\theta}(\vz_i,\vs),    \va\right)\right]$.}
            \parState{Calculate $ \mathcal{L}_{\mathrm{SOG}}=\hat\E_{\tau_E}\left[\mathcal{L}\left(f_{\boldsymbol\theta}(\vz_E,\vs), \va\right)\right]$ and its gradients w.r.t. $\boldsymbol\theta$.}
            \parState{Update $f_{\boldsymbol\theta}$ by stochastic gradient descent on $\boldsymbol\theta$ to minimize $\mathcal L_{\mathrm{SOG}}$.}
		\EndFor
	\end{algorithmic}
\end{algorithm}
\begin{algorithm}
	\caption{SOG-GAIL: Multimodal Combination of BC and GAIL}
	\label{alg:sog-gail}
	\begin{algorithmic}[1]
	    \State \textbf{Input:} Initial parameters of policy and discriminator networks, $\boldsymbol\theta_0, \vw_0$; expert trajectories $\tau_E\sim\pi_E$
		\For {$i=0,1,2,\ldots$}
		     \parState{Sample a latent code $\vz_i\sim p(\vz)$, and subsequently a trajectory $\tau_i\sim\pi_{\boldsymbol\theta_i}(\cdot|\vz_i)$.}
		     \parState{Sample state-action pairs $\chi_i\sim\tau_i$ and $\chi_E\sim\tau_E$ with the same batch size.}
		     \parState{Update the discriminator parameters from $\vw_i$
to $\vw_{i+1}$ with the gradient}
            \begin{equation*}
            \begin{split}
                &\mathbb{\hat{E}}_{\tau_i}\left[\nabla_{\vw}\log D_{\vw}(\vs,\va)\right]+
                \\&\mathbb{\hat{E}}_{\tau_E}\left[\nabla_{\vw}\log \left(1-D_{\vw}(\vs,\va)\right)\right].
            \end{split}
            \end{equation*}
            \parState{Calculate the surrogate loss $\mathcal{L}_\mathrm{PPO}$ using the PPO rule with the following objective}
            \begin{equation*}
                \mathbb{\hat{E}}_{\tau_i}\left[\nabla_{\vw}\log D_{\vw}(\vs,\va)\right]-\lambda_H H(\pi_{\boldsymbol\theta_i}).
            \end{equation*}
            \State Calculate the SOG loss $\mathcal{L}_{\mathrm{SOG}}$ per \Cref{alg:sog-bc}.
            \parState{Take a policy step from $\boldsymbol\theta_i$ to $\boldsymbol\theta_{i+1}$ w.r.t. the objective $\mathcal{L}_\mathrm{PPO}+\lambda_S \mathcal{L}_{\mathrm{SOG}}$.}
		\EndFor
	\end{algorithmic}
\end{algorithm} 
\section{Further Details of the Experiments}
In this appendix, we first present details about the setup of the experiments. Then, we demonstrate the robustness of our method against noise. Lastly, we show that a meaningful correspondence holds between the latent space and target locations in the Fetch experiment.
\subsection{Experiment Setup}
\label{appx:experiment-details}
In the Circles environment, the observed state at time $t$ is a concatenation of positions from time $t-4$ to $t$. Expert demonstrations appear in three modes, each trying to produce a distinct circle-like trajectory. The expert tries to maintain a $(2\pi/100)$ rad/s angular velocity along the perimeter of a circle. However, its displacement at each step incurs a 10$\%$-magnitude 2D Gaussian noise from the environment. 

To train an expert policy for MuJoCo locomotion tasks, we used Pearl \citep{rakelly2019efficient}, a few-shot reinforcement learning algorithm. For the FetchReach experiment, we considered a pretrained DDPG+HER model as expert \citep{lillicrap2015continuous,andrychowicz2017hindsight}.

In \Cref{sec:experiments}, we measured the rewards of different experiments as follows. For the Circles experiment, we applied a Gaussian kernel to the distance of the agent from the perimeter of each of the three reference trajectories. Hence, each step yields a reward between 0 and 1. For all MuJoCo locomotion environments, i.e. Ant, HalfCheetah, Walker2d, Hopper, and Humanoid, we used the original rewards introduced in \citet{rakelly2019efficient}. We ran all experiments with four random seeds and presented the results for the best training iteration in terms of expected sum of rewards. 

In experiment with varied velocities, having a suffix ``Vel'' in their name, target velocities of the expert are sampled uniformly from the values listed in \Cref{tab:possible-velocities}. As a reference, HalfCheetah has a torso length of around \hbox{1 m}.
\begin{table*}[!ht] 
\caption{Set of possible expert velocities in different environments.}
\label{tab:possible-velocities}
\begin{center}
\begin{scriptsize}
\begin{tabular}{lccc}
\toprule
Environment & Walker2d-Vel & Hopper-Vel & HalfCheetah-Vel\\
\midrule
Possible Velocities & $\{-1.5, -1 , -0.5,  0.5,  1 ,  1.5\}$ & $\{-2 , -1.5,  0.5,  1 ,  1.5,  2\}$ & Interval of $[1.5,3]$ \\
\bottomrule
\end{tabular}
\end{scriptsize}
\end{center}
\end{table*}

Episode lengths in different environments are listed in \Cref{tab:horizon}.
\begin{table*}[!ht] 
\caption{Episode lengths in different environments.}
\label{tab:horizon}
\begin{center}
\begin{scriptsize}
\begin{tabular}{lccccccc}
\toprule
Environment & Circles & Ant & HalfCheetah & Humanoid & Walker2d & Hopper & FetchReach\\
\midrule
Episode Length & $1000$ & $200$ & $200$ & $1000$ & $1000$ & $1000$ & $50$ \\
\bottomrule
\end{tabular}
\end{scriptsize}
\end{center}
\end{table*}

\begin{figure} 
    \centering
    \begin{subfigure}[t]{0.28\textwidth}
        \centering
        \includegraphics[height=1.5in]{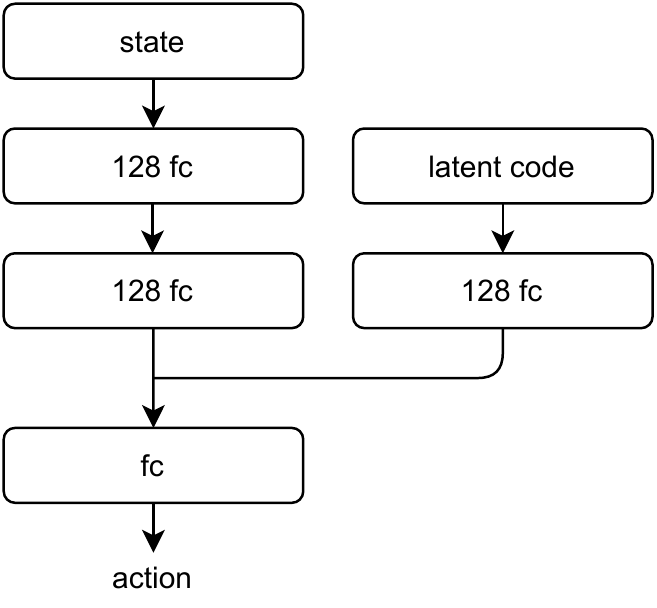}
        \caption{Policy network}
    \end{subfigure}
    \begin{subfigure}[t]{0.18\textwidth}
        \centering
        \includegraphics[height=1.5in]{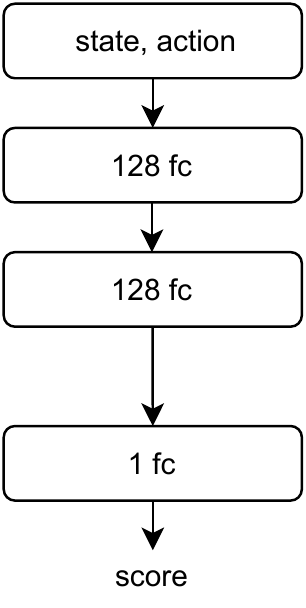}
        \caption{The discriminator}
    \end{subfigure}
    \caption{\label{fig:arch}\textbf{Neural network architectures.} (a) the policy network, (b) the discriminator in GAIL.}
\end{figure}

We used the same network architectures across all algorithms. The policy network passes the states and latent codes through separate fully connected layers, and then adds them and feeds them to the final layer. The discriminator receives state-action pairs and produces scores for the GAN objective. The posterior network for InfoGAIL has the same architecture as the discriminator, except it has a final softmax layer in the discrete case. For continuous latent codes, InfoGAIL adopts a Gaussian posterior. The architectures are illustrated in \Cref{fig:arch}.

For each of the algorithms SOG-GAIL, InfoGAIL, and VAE-GAIL, we pretrain the policy network with behavior cloning for better performance. Since the policy network has an input head for the latent code, in SOG-GAIL we provide the latent codes according to \Cref{alg:sog-bc}. Also, because InfoGAIL is not equipped with a module to extract the latent codes prior to training, we feed random samples as the input latent codes during pretraining. 

Across all experiments, we varied the coefficients $\lambda_I$, corresponding to the mutual information term in InfoGAIL, and $\lambda_S$ from SOG-GAIL (\Cref{alg:sog-gail}), at values of $\lambda=i*10^j;\ i=1,2,5;\ 0.01\leq\lambda\leq 1$.

\subsection{Robustness Analysis} \label{appx:robust}
We design two experiments to show that SOG-GAIL learns policies that are robust towards unseen states. In the first experiment, upon generating test trajectories for Ant-Dir-6, we switch the movement angles to all six possible directions. Since the motion angles within each expert trajectory are consistent, such changes create unseen circumstances. We observe that maneuvers with acute angles causes the agent trained with SOG-BC to topple over and make the episode terminate. In contrast, SOG-GAIL manages to switch between all six directions. 

\begin{figure}
\centering
\begin{minipage}{.43\textwidth}
  \centering
  \begin{subfigure}[b]{0.30\textwidth}
        \centering
        \adjincludegraphics[width=\textwidth,clip]{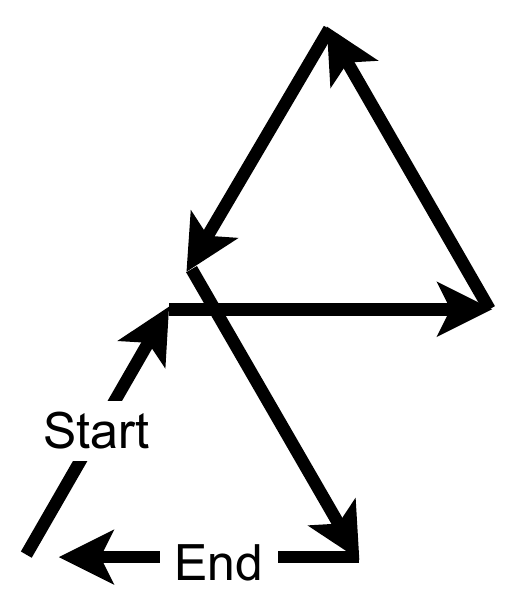}
        \caption{\label{fig:robust-goal}\small{Goal}}
    \end{subfigure}
    \begin{subfigure}[b]{0.33\textwidth}
        \centering
        \adjincludegraphics[width=\textwidth,trim={{.05\width} {.05\height} {.05\width} {.1\height}},clip]{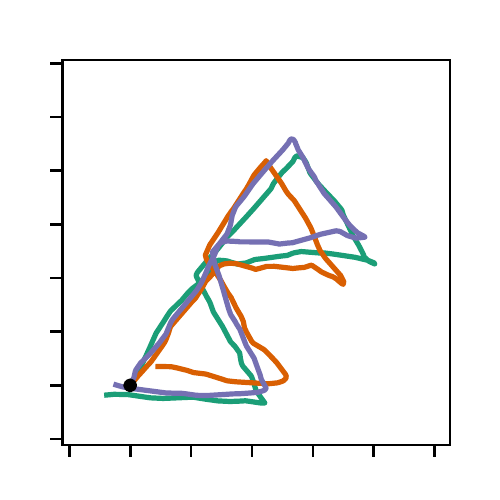}
        \caption{\small{SOG-GAIL}}
    \end{subfigure}
        \begin{subfigure}[b]{0.33\textwidth}
        \centering
        \adjincludegraphics[width=\textwidth,trim={{.05\width} {.05\height} {.05\width} {.1\height}},clip]{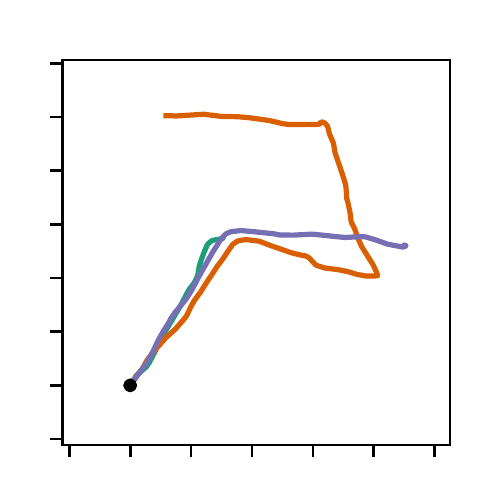}
        \caption{\label{fig:robust-SOG}\small{SOG-BC}}
    \end{subfigure}
    \caption{\label{fig:robustness} \textbf{Robustness of SOG-BC vs SOG-GAIL: Ant-Dir-6.} To create unseen situations, we switch the latent codes to those corresponding to the directions shown in (a). In (b), (c) we show three trajectories generated in different colors. We observe that the ant agent trained with SOG-GAIL performs the desired task flawlessly, while the one trained with SOG-BC often topples over and fails the task.}
\end{minipage}%
\hfill
\begin{minipage}{.97\linewidth}
  \centering
  \begin{subfigure}[b]{0.48\textwidth}
        \centering
        \adjincludegraphics[width=\textwidth,trim={{.00\width} {.00\height} {.1\width} {.1\height}},clip]{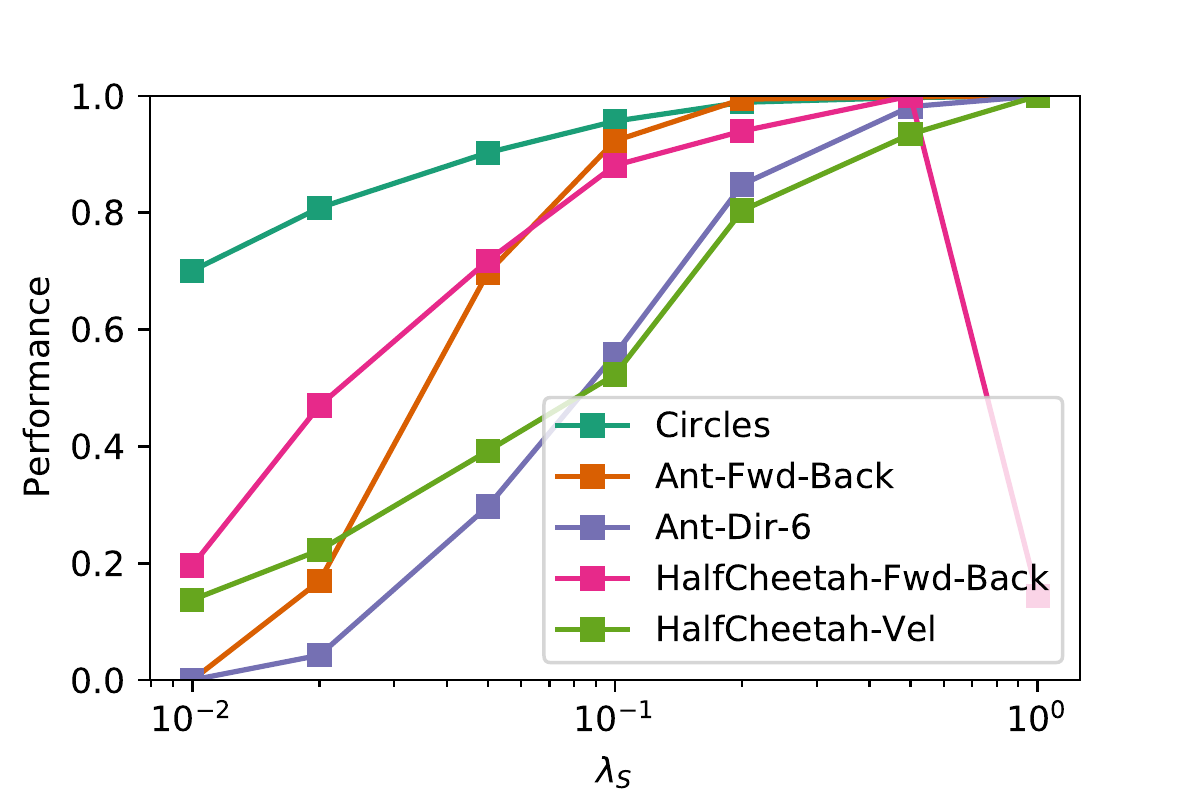}
        \caption{Unperturbed policy}
        \label{fig:unperturbed}
    \end{subfigure}
    \begin{subfigure}[b]{0.48\textwidth}
        \centering
        \adjincludegraphics[width=\textwidth,trim={{.00\width} {.00\height} {.1\width} {.1\height}},clip]{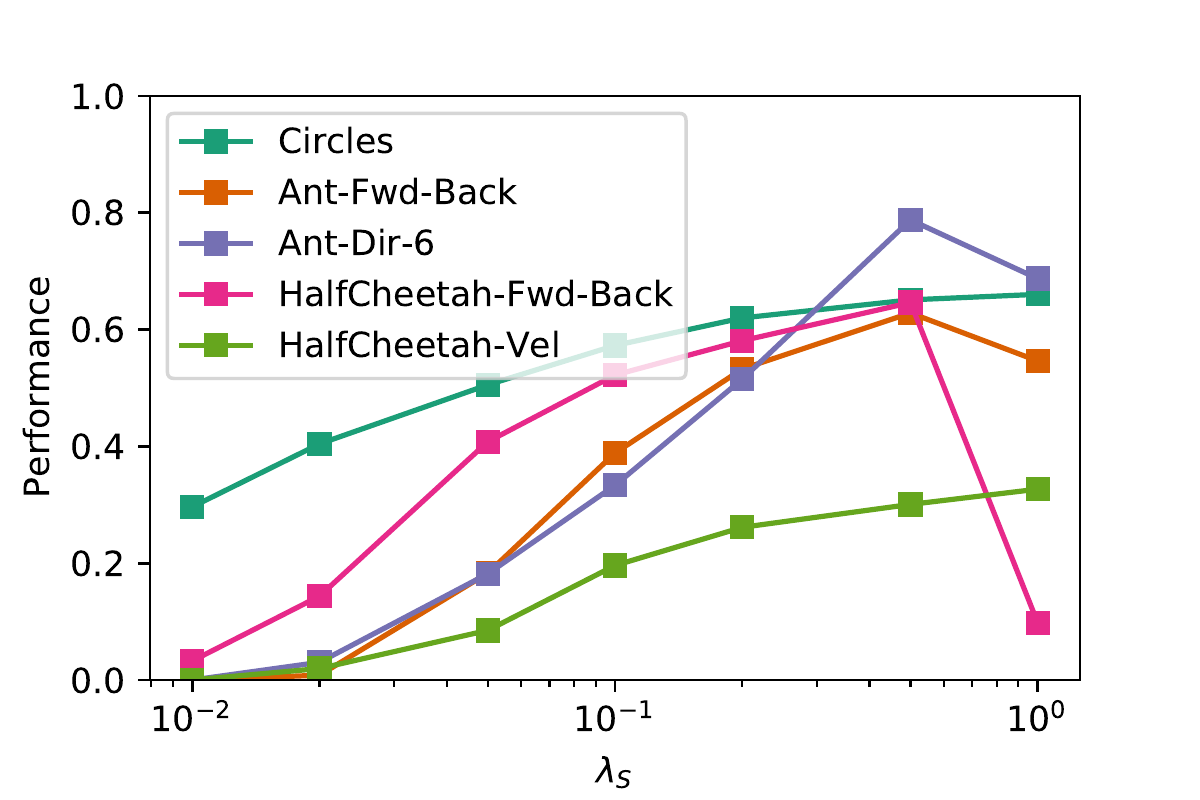}
        \caption{Perturbed policy}
        \label{fig:perturbed}
    \end{subfigure}
    \caption{\textbf{Optimal choice of $\bm{\lambda_S}$.} Parameter $\lambda_S$ is the coefficient of the SOG loss in \Cref{alg:sog-gail}. Performance of \Cref{alg:sog-gail} is illustrated as $\lambda_S$ varies. We consider two cases of with/without perturbations. These perturbations are imposed by taking random actions with a chance of 20\%. The vertical axis is normalized such that the best performance in unperturbed settings achieves a score of 1, and a random policy achieves 0. In Figure (b), we observe that the policy trained with higher values of $\lambda_S$ shows less robustness to perturbations in several experiments .}
  \label{fig:perturb}
\end{minipage}
\end{figure}

Secondly, we vary the coefficient $\lambda_S$ in SOG-GAIL in several experiments. High values of $\lambda_S$ in limit lead to SOG-BC. We observe that increasing $\lambda_S$ improves scores in \Cref{tab:rewards} across all experiments (except for HalfCheetah-Dir). We now perturb each learned policy by taking random actions with a 20\% probability. We observe that under this perturbation, several of the experiments perform worse at higher values of $\lambda_S$. This indicates that combination of SOG with GAIL is more robust to perturbations. The results of our two tests are respectively depicted in \Cref{fig:robustness,fig:perturb}.


\subsection{Latent Space Interpretability} \label{appx:latent-interpret}
\begin{figure*}[htp]
    \hspace{0.05\textwidth}
     \begin{subfigure}[b]{0.24\textwidth}
         \centering
         \adjincludegraphics[width=\textwidth,clip]{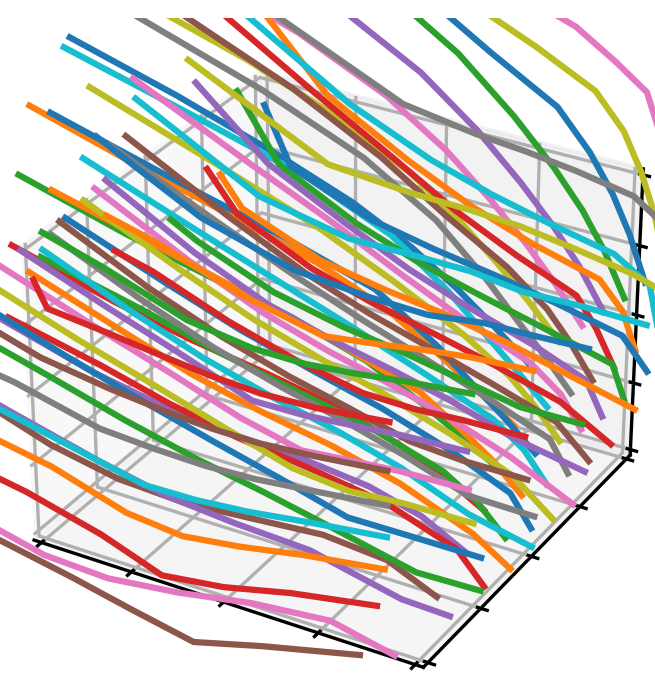}
     \end{subfigure}
     \hfill
     \begin{subfigure}[b]{0.24\textwidth}
         \centering
         \adjincludegraphics[width=\textwidth,clip]{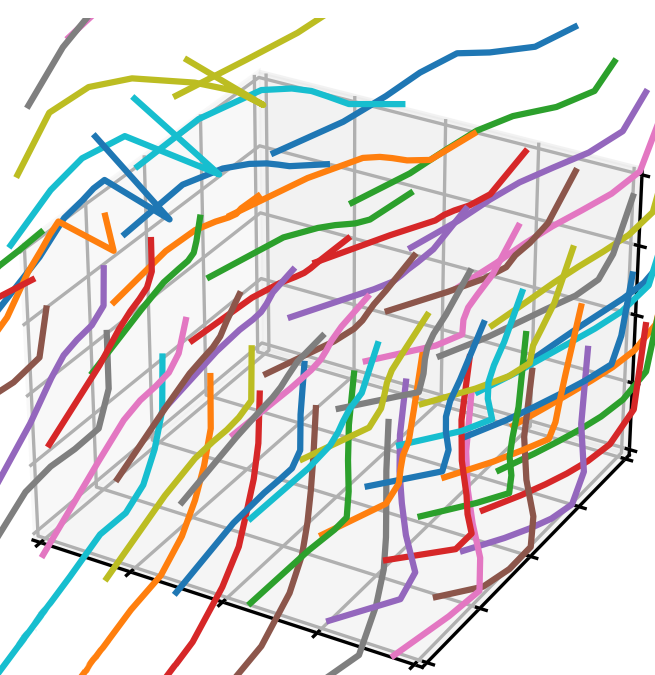}
     \end{subfigure}
     \hfill
     \begin{subfigure}[b]{0.24\textwidth}
         \centering
         \adjincludegraphics[width=\textwidth,clip]{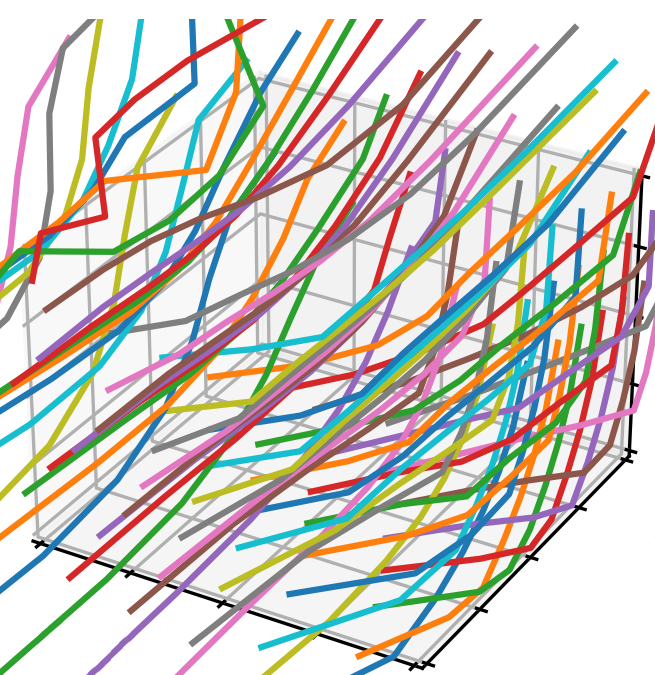}
     \end{subfigure}
    \hspace{0.05\textwidth}
        \caption{ \small \textbf{Self-organization in the latent space: FetchReach.} In each plot, two dimensions among the three dimensions of the latent space are fixed, whereas the third is varied. Each line corresponds to the location of the robotic arm at the terminal state of different trajectories. In each line, two latent dimensions take a fixed value, while the third varies. Lines are randomly colored for better visual distinction.
     }
\label{fig:fetch-latent}
\end{figure*}

In \Cref{fig:fetch-latent} we demonstrate that in the Fetch experiment, different areas of the latent space smoothly correspond to various target points.

\section{Optimization of GAIL} \label{appx:gail-optim}

GAIL optimizes the objective of \Cref{eq:gail} in two alternating steps: (1) a gradient ascent step to increase it with respect to the discriminator parameters, and (2) a Trust Region Policy Optimization (TRPO) \citep{schulman2015trust} or Proximal Policy Optimization (PPO) \citep{schulman2017proximal} step to decrease it with respect to the policy parameters. 

TRPO and PPO are policy optimization algorithms that take the biggest possible policy improvement step without stepping so far that performance accidentally deteriorates. In particular, PPO achieves this by optimizing the following ``surrogate'' objective at $i^{th}$ iteration:
\begin{equation} \label{eq:surrogate}
\begin{split}
&\mathcal{L}_\mathrm{PPO}(\vs,\va,\boldsymbol\theta_i,\boldsymbol\theta) = 
\\&\min\left(
\frac{\pi_{\boldsymbol\theta}(\va|\vs)}{\pi_{\boldsymbol\theta_i}(\va|\vs)}  A^{\pi_{\boldsymbol\theta_i}}(\vs,\va), \;\;
g(\epsilon, A^{\pi_{\boldsymbol\theta_i}}(\vs,\va))
\right),
\end{split}
\end{equation}
where $A$ is the advantage function \citep{schulman2015high}, and $g$ is a ``clipping'' function for some small parameter \hbox{$\epsilon>0$}:
\begin{equation}
g(\epsilon, A) = \left\{
    \begin{array}{ll}
    (1 + \epsilon) A & A \geq 0 \\
    (1 - \epsilon) A & A < 0.
    \end{array}
    \right.
\end{equation}
\section{Discrete Latent Variables: Maximum Marginal Likelihood Approach vs SOG} \label{appx:em}
Expectation-maximization (EM) is a well-known method for optimizing the maximum likelihood objective --marginalized over the latent variables -- in  models involving latent variable.  In this section, we demonstrate how the iterative procedure of \Cref{alg:sog} can be derived as a special case of the EM algorithm when the latent variable has a discrete distribution.
\subsection{The EM Algorithm in General}
First, we introduce the EM method in general. Let $\mX=\{\vx_i\}_{i=1}^{N}$, $\mY=\{\vy_i\}_{i=1}^{N}$, $\mZ=\{\vz_i\}_{i=1}^{N}$ constitute a dataset generated by the procedure described in \Cref{sec:method}, and let $\boldsymbol\Pi = \{\pi_{k}\}_{k=1}^{K}$ denote the set of probability masses for the discrete latent codes. The EM framework defines a total likelihood function for any given arbitrary distribution $q(\mZ)$ as follows:
\begin{subequations}
\begin{align}
    \begin{split}
    \LL(&q(\mZ); \boldsymbol\theta, \boldsymbol\Pi) \\=&
    \E_{\mZ \sim q(\mZ)} \log\left\{p(\mY, \mZ | \mX;\:\boldsymbol\theta, \boldsymbol\Pi)\right\}+ H(q(\mZ))\end{split}\\
    \begin{split}
    =& \E_{\mZ \sim q(\mZ)} \log\left\{p(\mY | \mX;\:\boldsymbol\theta, \boldsymbol\Pi)\right\} \\&+ \E_{\mZ \sim q(\mZ)} \log\left\{ \frac{p(\mZ | \mX,\mY; \:\boldsymbol\theta, \boldsymbol\Pi)}{q(\mZ)} \right\}
    \end{split}
    \\
    \begin{split}
    \equiv& \log p(\text{\texttt{data}} | \boldsymbol\theta, \boldsymbol\Pi)\\& - \KL\big({q(\mZ)}\:||\:{p(\mZ | \mX,\mY;\: \boldsymbol\theta, \boldsymbol\Pi)}\big)
    \end{split}
\end{align}
\end{subequations}
After initialization of model parameters, the optimization proceeds by alternating between two steps:\\
\begin{enumerate}
    \item \textbf{Expectation Step (E Step).} 
    \begin{subequations}
    \begin{align}
        q^{t+1}(\mZ) &\leftarrow \displaystyle \argmax_q \LL(q;\:\boldsymbol\theta^{t}, \boldsymbol\Pi^{t}) \\&=  p(\mZ | \mX,\mY;\: \boldsymbol\theta^{t}, \boldsymbol\Pi^{t}) \\&\equiv \prod_{i=1}^N r_{i,\vz_{i}}^{t}
    \end{align}
    \end{subequations}
\item \textbf{Maximization Step (M Step).} 
\begin{subequations}
\label{eq:M-step}
\begin{align}
 \begin{split}
 \boldsymbol\theta^{t+1}, &\boldsymbol\Pi^{t+1} \\ \leftarrow&\argmax_{\boldsymbol\theta, \boldsymbol\Pi}\LL(q^{t+1};\:\boldsymbol\theta, \boldsymbol\Pi)
 \end{split}\\
 =\;& {\argmax_{\boldsymbol\theta, \boldsymbol\Pi}\: \mathbb{E}_{\mZ\sim q^{t+1}}[\log P(\mY,\mZ | \mX;\: \boldsymbol\theta, \boldsymbol\Pi)]} \\
 \begin{split}
 =\;& \argmax_{\boldsymbol\theta, \boldsymbol\Pi} \sum_{i=1}^N \sum_{k=1}^{K} r_{ik}^{t}(\log \pi_{k} \\&+ \log p(\vy_i | \vx_i, \vz_i=k;\: \boldsymbol\theta))
 \end{split}
 \label{eq:soft-em-mstep}
\end{align}
\end{subequations}
\end{enumerate}
where 
\begin{equation}
\begin{split}
r_{ik}^{t} &= p(\vz_i=k \given \vx_i, \vy_i;\: \boldsymbol\theta^{t}, \boldsymbol\Pi^{t}) \\&= \frac{p(\vy_i | \vx_i,\vz_i=k;\: \boldsymbol\theta^{t})\:\pi_k^t}{\sum_{l=1}^K p(\vy_i | \vx_i, \vz_i=l; \:\boldsymbol\theta^{t})\: \pi_l^t}.
\label{eq:em-post}
\end{split}
\end{equation}
Finally, we introduce the update rules in the M step for $\boldsymbol\theta,\boldsymbol\Pi$. Since the M step does not have a closed-form solution w.r.t. $\boldsymbol\theta$, on can instead perform a batch gradient ascent step. This approach for an intractable M step is known as generalized EM \citep{bishop2006pattern}. Also, it is straightforward to derive the following update rule for $\boldsymbol\Pi$ by maximizing the objective of \Cref{eq:soft-em-mstep} subject to the constraint $\sum_{k=1}^K\pi_k=1$:
\begin{equation} \label{eq:prior-update}
    \pi_k^t \leftarrow \frac{\sum_i r_{ik}^t}{\sum_{il} r_{il}^t}.
\end{equation}

The E step can be carried out in two ways:
\begin{enumerate}[1.]
    \item \textbf{Soft.} using all the posterior probabilities in \Cref{eq:em-post}, so that each data point has a soft-distribution over the latent codes at each iteration step. This corresponds to the original EM algorithm above.
    \item \textbf{Hard.} approximating the posterior probabilities by a one-hot distribution, driven by the best latent code for the M step. In the following subsection, we derive when the regular EM reduces to the Hard EM method.
\end{enumerate}

\subsection{Hard Assignment EM and SOG (\Cref{alg:sog})}
Hard assignment EM has a long history in the ML community and the assumptions made to derive this algorithm are well known and are used to motivate and justify the $K$-means algorithm from an EM perspective \citep{bishop2006pattern}. 
As in SOG, one replaces the probabilities $r_{ik}$ in \Cref{eq:em-post} with binary values denoted as $\gamma_{ik}$. This hard assignment is performed by allocating all the probability mass to the categorical variable for which the posterior probability is the maximum. 

Let $k_i^* = \argmax_{k} p(\vz_i=k \given \vx_i, \vy_i;\: \boldsymbol\theta^{t}, \boldsymbol\Pi^{t})$. Then in the Hard EM case, the E step becomes
\begin{equation}\label{eq:em-post-1hot}
\gamma_{ik}^{t} = 
\begin{cases}1, &  k = k_i^* \; 
\\ 0, & k\not= k_i^*
\end{cases},
\end{equation}
and the M step of \Cref{eq:soft-em-mstep} (w.r.t. $\boldsymbol\theta$) simplifies to:
\begin{equation} \label{eq:sog-mstep}
    \argmin_{\boldsymbol\theta}\sum_{i=1}^N ||f(\vz_i=k_i^*,\vx_i;\boldsymbol\theta)-\vy_i||^2.
\end{equation}
It follows from  \Cref{eq:em-post} that if $\pi_k=1/K;\:k=1,\ldots, K$, then
\begin{equation*}
\begin{split}
    &\argmax_{k\in\{1,\ldots,K\}}p(\vz_i=k \given \vx_i, \vy_i;\: \boldsymbol\theta^{t}) \\&\hspace{.3in} =\argmax_{k\in\{1,\ldots,K\}}p(\vy_i \given \vz_i=k, \vx_i, ;\: \boldsymbol\theta^{t}).
\end{split}
\end{equation*}
Hence, the Hard EM is equivalent to the update step in \Cref{alg:sog} in this special case.

Next we provide a sufficient condition under which  both the Hard EM and SOG (for any values of $\pi_k$'s)  become a special case of the general  EM.   First, recall that
\begin{equation}
\begin{split}
&p(\vy_i|\vz_i=k,\vx_i;\,\boldsymbol\theta)\\&\quad\propto\exp[-||f(\vz_i=k,\vx_i;\,\boldsymbol\theta)-\vy_i||^2/(2\sigma^2)],
\end{split}
\end{equation}
where $\sigma$ determines the spread around the manifold of data. Let 
\begin{subequations} 
\begin{align}
    k_i^\dagger &= \argmax_{k_i\in\{1,\ldots,K\}}p(\vy_i|\vz_i=k,\vx_i;\,\boldsymbol\theta)\\
    &=\argmin_{k_i\in\{1,\ldots,K\}} ||f(\vz_i=k,\vx_i;\boldsymbol\theta)-\vy_i||^2.
\end{align}
\end{subequations}
Assuming a very small spread $\sigma$,  we get for all $k\not= k_i^\dagger$ 
\begin{equation}
    \lim_{\sigma \rightarrow 0} \frac{p(\vy_i|\vz_i=k, \vx_i; \,\boldsymbol\theta)}{p(\vy_i|\vz_i= k_i^\dagger,\vx_i; \,\boldsymbol\theta)} = 0\; ,
\end{equation}
and substituting these probabilities in the equation for $r_{ik}^t$ in  \Cref{eq:em-post},
we obtain that the posteriors follow the same pattern:
\begin{equation}
\lim_{\sigma \rightarrow 0} p(\vz_i=k|\vx_i, \vy_i;\,\boldsymbol\theta) = 
\begin{cases}1, &  k = k_i^\dagger \; 
\\ 0, & k\not= k_i^\dagger
\end{cases},
\end{equation}

Thus, the SOG algorithm that computes  an MLE estimate of $\vz$ at every iteration (i.e. the value of the latent variable for which  data likelihood is maximized given $\boldsymbol\theta$) is equivalent to the standard EM when $\sigma \approx 0$. 
\section{Continuous Latent Variables: Maximum Marginal Likelihood Approach vs SOG} \label{appx:continuous}
In this section, we consider the case of continuous latent variables, and demonstrate that in the asymptotic limit where $\sigma$ is sufficiently small, the same relationship derived in the preceding section for the discrete case still holds:  The marginalized data log-likelihood $p(\vy|\vx;\,\boldsymbol\theta)$  is well approximated by the maximum of the data log-likelihood computed over the latent variables as done in  \Cref{alg:sog}. In this limit, we can use Laplace approximation for integrals. Laplace approximation provides a general way to approach marginalization problems. The basic setting for Laplace approximation is a multivariate integral of the form:
\begin{equation} \label{eq:laplace-main}
    I=\int e^{\,-h(\vz)/t}\:d\vz,
\end{equation}
where $\vz$ is a $d$-dimensional variable, and $h(\vz)$ is a scalar function with a unique global minimum at $\vz=\vz^*$.
Laplace approximation considers a Taylor series expansion of $h(\vz)$ around $\vz^*$. In this way, as $t\rightarrow 0\,$, the integral of \Cref{eq:laplace-main} is approximated as:
\begin{equation} \label{eq:laplace-approx}
    I= e^{\,-h(\vz^*)/t}\:(2\pi)^{d/2}\:|\mA|^{1/2}\:t^{\,d/2}\ +O(t),
\end{equation}
where $\mA=(\mD^2 h(\vz^*))^{-1}$ is the inverse of the Hessian of $h(\vz)$ evaluated at $\vz^*$.

Now, consider the expression for the data likelihood of the model in \Cref{sec:method}
\begin{subequations}
\begin{align}
    I&=p(\vy|\vx;\, \boldsymbol\theta) \\
    &=\int p(\vy|\vz, \vx;\, \boldsymbol\theta)p(\vz)d\vz\\
    \begin{split}
    &\propto \int\exp\left[-||f(\vz,\vx)-\vy||^2/(2\sigma^2)\right] \\
    &\hspace{.3in}\times\exp\left[-||\vz||^2/2\right]d\vz. \\
    \end{split}
\end{align}
\end{subequations}
For $\sigma$ sufficiently small, we can use Laplace approximation for this integral. Specifically, comparing with \Cref{eq:laplace-main} we identify:
\begin{subequations}\label{eq:laplace-term-matching}
\begin{align}
        h(\vz)&\coloneqq||f(\vz,\vx)-\vy||^2+\sigma^2||\vz||^2,\\
        t&\coloneqq 2\sigma^2,
\end{align}
\end{subequations}
thus deriving:
\begin{subequations}
\begin{align}
    \vz^*&=\argmin h(\vz)\\
    &=\argmin||f(\vz,\vx)-\vy||^2+\sigma^2||\vz||^2\\
    &\approx \argmin||f(\vz,\vx)-\vy||^2\label{eq:laplace-z-choice},
\end{align}
\end{subequations}
which is equivalent to \cref{alg:sog-best-z} of \Cref{alg:sog}. The only nuance is that \Cref{alg:sog} finds an approximation to $\vz^*$ by searching among a batch of sampled candidate $\vz$'s. However, given large batches of $\vz$'s, the best candidate stays close to $\vz^*$ with a high probability. Sampling of $\vz$ from a Gaussian prior ensures that $||\vz||$ remains small, making the approximation of \Cref{eq:laplace-z-choice} valid.

Plugging the terms in \Cref{eq:laplace-term-matching} into \Cref{eq:laplace-approx}, we obtain the following approximation to the data log-likelihood, which is later maximized w.r.t. model parameters $\boldsymbol\theta$, 
\begin{equation} \label{eq:laplace-log-likelihood}
   \begin{split}
       \log p(\vy|\vx;\, \boldsymbol\theta)\approx-\,&\frac{1}{2\sigma^2}\left(||f(\vz^*,\vx)-\vy||^2+\sigma^2||\vz^*||\right)\\
       +&\,\frac{1}{2}\log|\mA|\\
       +&\,d\log\sigma\\
       +&\,\text{constant}.
   \end{split} 
\end{equation}

We notice that in the asymptote of $\sigma\rightarrow 0$, the term $||f(\vz^*,\vx)-\vy||^2\,/(2\sigma^2)$ dominates the other terms. Particularly, $||\vz^*||$ and $\log|\mA|$ do not depend on $\sigma$, and the growth of $\log\sigma$ is dominated by that of $1/\sigma^2$. Therefore, in the limit we can approximate \Cref{eq:laplace-log-likelihood} with only the first summand. That is,
\begin{equation} \label{eq:laplace-log-likelihood-approx}
       \log p(\vy|\vx;\, \boldsymbol\theta)\approx-\,\frac{1}{2\sigma^2}\,||f(\vz^*,\vx)-\vy||^2,
\end{equation}
which yields \cref{alg:sog-loss} of \Cref{alg:sog} since $\sigma$ is constant w.r.t. $\boldsymbol\theta$. We assume Lipchitz continuity for $f$. Thus, the approximate choice of $\vz^*$, given large batches over the latent codes, introduces an insignificant error to \Cref{eq:laplace-log-likelihood-approx}.

It is worth mentioning that similar approximation results are well studied in the framework of Bayesian model selection. In that context, the data likelihood is  marginalized instead over the model parameters, and also the asymptotic variable is rather the number of observed data samples becoming large  \cite{bishop2006pattern}.
\section{\label{appx:lipschitz-bounds}The Self-Organization Effect}
In this section, we discuss how the  self-organization phenomenon, that is the formation of similar output data in different regions of the latent space, naturally emerges in \Cref{alg:sog}. In the following, we present three propositions demonstrating different aspects of the self-organization phenomenon in the latent space, under the assumption that the function $f(\cdot)$  is a sufficiently smooth function of its inputs $(\vx,\vz)$, e.g. it is Lipschitz continuous:
\begin{enumerate}
\item \textit{Continuity in the data space:} We discuss that for every data point, there is a neighborhood in the data space such that for the same latent code $\vz$, the variations in the $L_2$-loss in \Cref{alg:sog} over all points in the neighborhood is small.
\item \textit{Two sufficiently close data points will have nearby winning latent codes (or the same winning latent codes in the discrete case):}  We show that if for a data point, a particular latent code results in a significantly better $L_2$-loss than another latent code, then the same  will also hold within a neighborhood of that data point. Thus, if $\vz$ is a winning code for a data point, then the points in its neighborhood will also have winning codes close to $\vz$  (or the same winning code $\vz$ in the discrete case). 
\item \textit{Network training can be a collaborative process:} We demonstrate that if we update the network parameters $\boldsymbol\theta$ to improve the $L_2$-loss for a particular data point (at a latent code $z$), then the same update will also collaboratively improve the loss for other data points within the neighborhood of the main data point (w.r.t. the same latent code $z$). 
\end{enumerate}
All three propositions provide insights on why the SOG training algorithm converges fast. In particular, in the mini-batch optimization in any implementation of the SOG algorithm, these propositions imply that the parameter updates for a sample batch of data points, reduce the loss for points neighboring to the sampled data points as well. 

Formally, we shall make the assumption that the network $f$ is a Lipschitz continuous function with the constant $\lpz$. We also define the $\delta_x$-$\delta_y$-neighborhood of a data point $(\vx,\vy)$ as $B(\vx_1,\vy_1;\delta_x,\delta_y) = \{(\vx,\vy) \given \norm{\vx - \vx_1} < \delta_x/\lpz,\: \norm{\vy- \vy_1} < \delta_y\}$. For later convenience, we define $\delta = \delta_x+\delta_y$. Finally, we denote $d(\vx,\vy,\vz;\boldsymbol\theta)\coloneqq\norm{f(\vx,\vz;\boldsymbol\theta)-\vy)}$.\vspace{.1in}

\begin{proposition} \label{prop:1}
For any data point $(\vx_1,\vy_1)$, if $(\vx_2,\vy_2) \in B(\vx_1,\vy_1;\delta_x,\delta_y)$, then $|d(\vx_1, \vy_1, \vz; \theta) - d(\vx_2, \vy_2, \vz; \theta)| < \delta_x+\delta_y=\delta$
\end{proposition}

\begin{proof}
From $(\vx_2, \vy_2) \in B(\vx_1, \vy_1;\delta_x,\delta_y)$, follows that  $\norm {\vx_1 - \vx_2} < \delta_x/\lpz$, $\norm {\vy_1 - \vy_2} < \delta_y$. Therefore: 
\begin{subequations}
\begin{align}
&|\norm{f(\vx_1,\vz;\boldsymbol\theta) - \vy_1} - \norm{f(\vx_2,\vz;\boldsymbol\theta) - \vy_2}| \\&\leq \norm{f(\vx_1,\vz;\boldsymbol\theta) - \vy_1 - ( f(\vx_2,\vz;\boldsymbol\theta) - \vy_2)} \label{eq:rev-tri}\\ 
&\leq \norm{f(\vx_1,\vz;\boldsymbol\theta) - f(\vx_2,\vz;\boldsymbol\theta)}+\norm{\vy_1 -\vy_2} \label{eq:tri}\\ 
& \leq \lpz \norm{\vx_1 - \vx_2}+\norm{\vy_1 -\vy_2} \label{eq:lip} \\ 
&< \delta_x + \delta_y = \delta.
\end{align}
\end{subequations}
where \Cref{eq:rev-tri,eq:tri} follow reverse triangle inequality and triangle inequality, respectively. Besides, \Cref{eq:lip} follows from the Lipschitz continuity assumption.
\end{proof}

\begin{proposition} \label{prop:2}
Let $(\vx_1,\vy_1)$ be a data point, and $\vz_i,\vz_j$ be two choices of the latent code.\\
If $\dfrac{d(\vx_1, \vy_1, \vz_i; \boldsymbol\theta)}{d(\vx_1, \vy_1, \vz_j; \boldsymbol\theta)} > C > 1$ for a constant $C$, and $d(\vx_1, \vy_1, \vz_j; \boldsymbol\theta)$ is finite, then $\exists \delta_x, \delta_y$ such that \[\forall (\vx_2,\vy_2) \in B(\vx_1, \vy_1;\delta_x,\delta_y),\quad \frac{d(\vx_2, \vy_2, \vz_i; \boldsymbol\theta)}{d(\vx_2, \vy_2, \vz_j; \boldsymbol\theta)}>1\]

\end{proposition}
\begin{proof}
$\forall \eta > 0$, $\exists \delta_x, \delta_y$ such that $\delta_x + \delta_y = \delta < \eta d(\vx_1, \vy_1, \vz_j; \boldsymbol\theta)$. Then using \cref{prop:1}, the following sequence of inequalities holds
\begin{subequations}
\begin{alignts}
d(\vx_2, \vy_2, \vz_i) &\geq d(\vx_1, \vy_1, \vz_i) - \delta \\
&> Cd(\vx_1,\vy_1, \vz_j) - \delta \\
&> Cd(\vx_1,\vy_1, \vz_j) - \eta d(\vx_1,\vy_1, \vz_j)\\
&= (C-\eta)d(\vx_1,\vy_1, \vz_j)\\
&> (C-\eta)(d(\vx_2,\vy_2, \vz_j) - \delta)\label{eq:prop:2:e}\\
&> (C-\eta)\left(1-\frac{\eta}{1-\eta}\right)d(\vx_2,\vy_2, \vz_j),\label{eq:prop:2:f}
\end{alignts}
\end{subequations}
where the step from (\ref{eq:prop:2:e}) to (\ref{eq:prop:2:f}) is done by realizing the fact that \[\eta d(\vx_2,\vy_2, \vz_j) > \eta \big(d(\vx_1,\vy_1, \vz_j) - \delta\big) > (1-\eta) \delta\] and therefore $\delta < \frac{\eta}{1-\eta} d(\vx_2,\vy_2, \vz_j)$.

It can be easily verified that there always exists a small enough value of $\eta\rightarrow 0^+$ such that \[(C-\eta)\left(1-\frac{\eta}{1-\eta}\right)>1\]
and therefore 
\[ d(\vx_2, \vy_2, \vz_i) > d(\vx_2,\vy_2, \vz_j). \]
\end{proof}
\begin{proposition} \label{prop:3}
Let $(\vx_1,\vy_1)$ be a data point and let $\vz$ be a choice of the latent code, respectively. Also assume an improvement update of the network parameters $\boldsymbol\theta_1$ to $\boldsymbol\theta_2$ that ensures $d(\vx_1, \vy_1, \vz; \boldsymbol\theta_1) > d(\vx_1, \vy_1, \vz; \boldsymbol\theta_2)$. Then, there exist $\delta_x$ and $\delta_y$ such that
\resizebox{0.95\linewidth}{!}{$
\forall (\vx_2, \vy_2) \in B(\vx_1, \vy_1;\delta_x, \delta_y),\;\;
d(\vx_2, \vy_2, \vz; \boldsymbol\theta_2) < d_{(\delta_x, \delta_y)}^{\;\max}(\boldsymbol\theta_1)$}, where
\[
    d_{(\delta_x, \delta_y)}^{\;\max}(\boldsymbol\theta) = \max_{(\vx, \vy) \in B(\vx_1, \vy_1;\delta_x, \delta_y)} d(\vx, \vy, \vz;\boldsymbol\theta )
\]
is the largest reconstruction distance for data points in the neighborhood of $(\vx_1, \vy_1)$
\end{proposition}
\begin{proof}
Let $a = d(\vx_1, \vy_1, \vz; \boldsymbol\theta_1) - d(\vx_1, \vy_1, \vz; \boldsymbol\theta_2) > 0$, and again $\delta = \delta_x + \delta_y$, then $\forall \delta_x, \delta_y$, and $\forall (\vx_2,\vy_2) \in B(\vx_1, \vy_1;\delta_x,\delta_y)$
\begin{subequations}
\begin{align}
d(\vx_2, \vy_2,\vz; \boldsymbol\theta_2) 
&\leq  d_{(\delta_x, \delta_y)}^{\;\max}(\boldsymbol\theta_2)\\
&\leq d(\vx_1, \vy_1, \vz; \boldsymbol\theta_2)+ \delta \label{eq:apply-dmax}\\
&\leq d(\vx_1, \vy_1, \vz; \boldsymbol\theta_1) - a + \delta\\
&\leq d_{(\delta_x, \delta_y)}^{\;\max}(\boldsymbol\theta_1) - a + \delta 
\end{align}
\end{subequations}
where \Cref{eq:apply-dmax} follows from \Cref{prop:1}.

Since $a$ is finite, $\exists \delta_x, \delta_y$ such that $\delta < a$, and hence
\[
d(\vx_2, \vy_2,\vz; \boldsymbol\theta_2) < d_{(\delta_x, \delta_y)}^{\;\max}(\boldsymbol\theta_1).
\]
\end{proof}

\section{Extension of Algorithm \ref{alg:sog} to Large Latent Spaces} \label{appx:extended-sog}

Algorithm \ref{alg:sog} suffers from an exponential growth of the number of samples required, as the latent space dimension increases. To address this, one can consider a coordinate-wise search of the latent code $\vz$, which results in \Cref{alg:sog-coordinate}. This modification reduces the computational cost from exponential to linear. 

\begin{algorithm}
	\caption{Coordinate-Wise Search in SOG}
	\label{alg:sog-coordinate}
	\begin{algorithmic}[1]
	    \parState{\textbf{Define:} Loss function $\mathcal{L}(\hat{\vy},\vy)\coloneqq ||\hat{\vy}-\vy||^2$, block size $\Delta$.}
		\For {$epoch=1,2,\ldots$}
		     \For {$iteration=1,2,\ldots$}
		        \parState{Sample a minibatch of $N_\text{data}$ data points $(\vx_i,\vy_i)$ from the dataset.}
		        \For {$i=1,2,\ldots,N_\text{data}$}
                    \State Initialize latent code $\vz_i=\boldsymbol{0}\in \mathbb{R}^d$
    		        \For {$b=1,2,\ldots,(d/\Delta)$}
    		            \parState{Initialize candidate latent codes\\
    		            \nonumber$\vz_{j}\coloneqq \vz_i;\ j=1,\ldots,N_z$.}
    		            \parState{Replace the $b$'th $\Delta$-element block in $\vz_{j}$ with a sample from $\mathcal{N} (\boldsymbol{0},\mI^{\Delta\times \Delta})$;\\
    		            \nonumber$\:j=1,\ldots,N_z$.}
    		            \parState{Calculate \\
    		            \nonumber $\vz_i\coloneqq\arg\min_{\vz_{j}}\mathcal{L}(f_{\boldsymbol\theta}(\vz_{j}, \vx_i), \vy_i)$.}
    	           \EndFor
    	       \EndFor
    	       \parState{Calculate $\mathcal{L}_{\mathrm{SOG}}=\sum_{i=1}^{N_\text{data}}\mathcal{L}(f_{\boldsymbol\theta}(\vz_j, \vx_i), \vy_i)$ and its gradient w.r.t. $\boldsymbol\theta$.}
    	       \parState{Update $f_{\boldsymbol\theta}$ by stochastic gradient descent on $\boldsymbol\theta$ to minimize $\mathcal L_{\mathrm{SOG}}$.}
		    \EndFor
		\EndFor
	\end{algorithmic}
\end{algorithm} 

\begin{figure}
    \centering
    \begin{subfigure}[t]{0.23\textwidth}
        \centering
        \parbox[b][1.1in][c]{0.5\hsize}{%
        \includegraphics[height=1.1in]{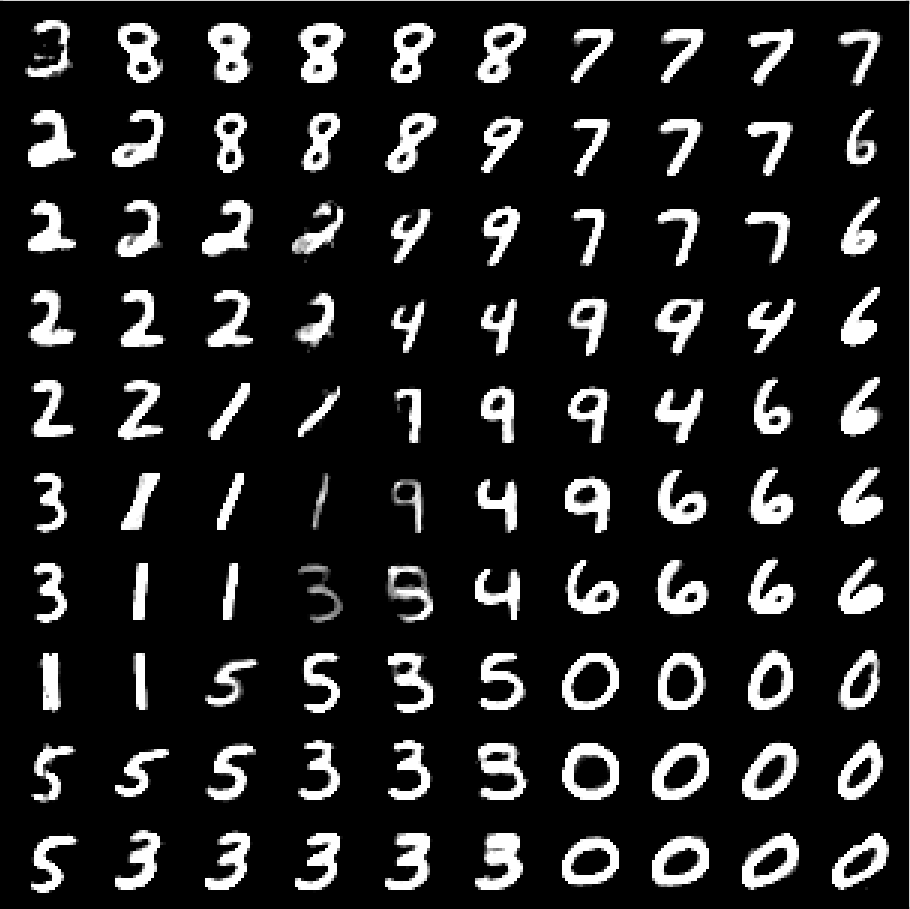}}
        \caption{Synthesized MNIST digits}
    \end{subfigure}
    \begin{subfigure}[t]{0.23\textwidth}
        \centering
        \includegraphics[height=1.1in]{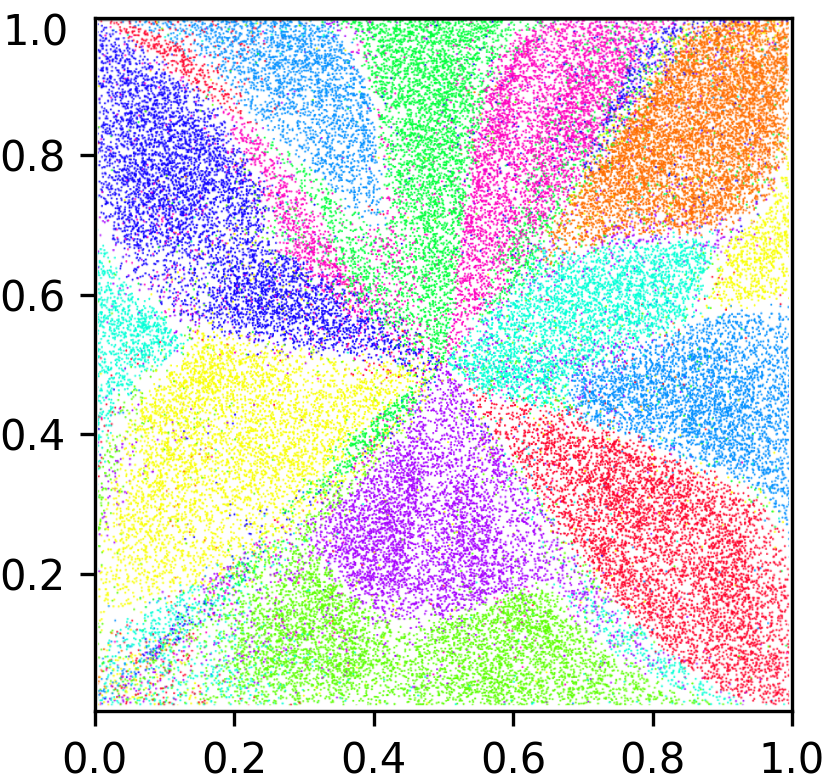}
        \caption{Embedding of MNIST test data in 2 dimensions}
    \end{subfigure}
    \caption{\label{fig:mnist}\textbf{MNIST dataset.} Output results of \Cref{alg:sog-coordinate} for the latent dimension of 2. (a) Synthesized images (b) Embedding of MNIST test data in 2 dimensions by the search mechanism of the SOG algorithm. Color coding represents ground truth digit labels. It is evident that despite the unsupervised training of the SOG algorithm, data samples corresponding to each digits are evenly organized across the latent space.}
\end{figure}

\begin{figure}
    \centering
    \includegraphics[width=0.95\hsize]{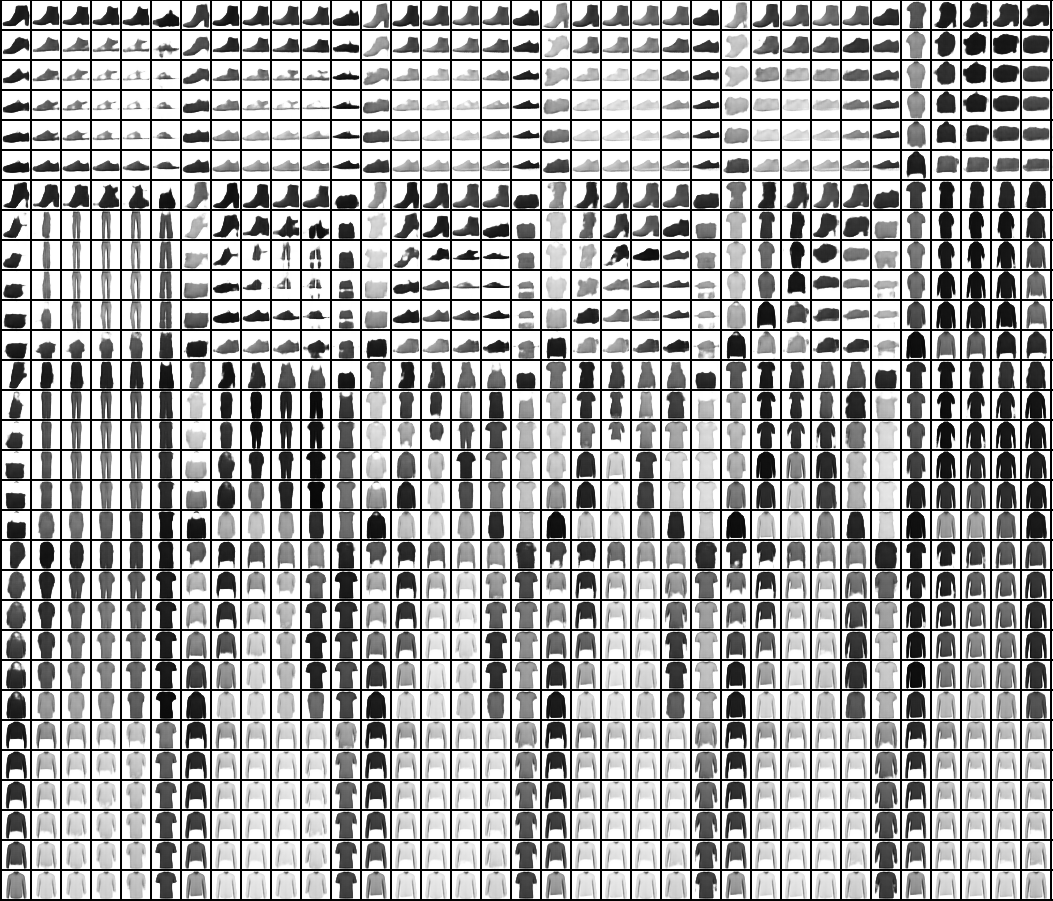}
    \caption{\label{fig:fashion-mnist}\textbf{Fashion MNIST.} Output results of \Cref{alg:sog-coordinate} for the latent dimension of 8. Each $6\times6$ block (or block of blocks, etc.) sweeps over one dimension of the latent space. Due to limited space, only part of the entire latent space is plotted. Full results are provided in the \underline{link} (see the summplementary material zip file in the review stage). Different latent dimensions have smoothly captured semantically meaningful aspects of the data, i.e. type of clothing, color intensity, thickness, etc.}
\end{figure}

\begin{figure} 
    \centering
    \begin{subfigure}[t]{0.23\textwidth}
        \centering
        \includegraphics[height=3in]{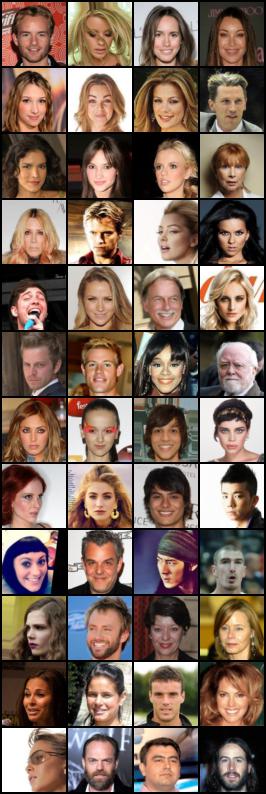}
        \caption{CelebA ground-truth data}
    \end{subfigure}
    \begin{subfigure}[t]{0.23\textwidth}
        \centering
        \includegraphics[height=3in]{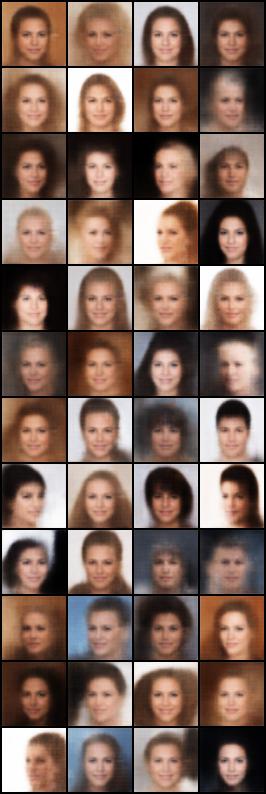}
        \caption{Reconstruction by SOG algorithm}
    \end{subfigure}
    \caption{\label{fig:celeba}\textbf{CelebA.} Output results of \Cref{alg:sog-coordinate} for the latent dimension of 16. (a) Ground truth samples. (b) Reconstructed samples synthesized by SOG. Note that the reconstructed images learned many of the salient features of the original faces, including the pose of the faces, as well as the hairstyle in most cases.  }
\end{figure}

In the FetchReach experiment, the latent dimension is 3 and we  adopted \Cref{alg:sog-coordinate} that resulted in superior results (see \Cref{fig:fetch}).  Besides, in \Cref{fig:mnist,fig:fashion-mnist,fig:celeba}, we present visual results of \Cref{alg:sog-coordinate} demonstrating the capability of our method in fitting complex distributions. The axes in \Cref{fig:mnist,fig:fashion-mnist} are the CDFs of the Gaussian latent codes for better visualization. We used the network architecture of the generator in \citet{radford2015unsupervised}.

\end{document}